\documentclass[]{fairmeta}
\usepackage{amsmath}
\usepackage{pifont}
\usepackage{listings}
\tcbuselibrary{breakable,listings}
\setcitestyle{square,comma,numbers,sort&compress}
\hypersetup{
    colorlinks,
    breaklinks,
    urlcolor = black,
    linkcolor = blue,
    citecolor = blue,
}

\lstdefinestyle{promptlisting}{
  basicstyle=\ttfamily\scriptsize,
  breaklines=true,
  breakautoindent=false,
  breakindent=0pt,
  columns=fullflexible,
  keepspaces=true,
  showstringspaces=false,
  xleftmargin=0pt,
  aboveskip=0pt,
  belowskip=0pt,
}
\tcbset{promptbox/.style={
  enhanced jigsaw,
  listing style=promptlisting,
  colback=black!2,
  colframe=black!55,
  colbacktitle=black!55,
  coltitle=white,
  fonttitle=\small,
  boxrule=0.6pt,
  arc=2pt,
  left=3pt, right=3pt, top=3pt, bottom=3pt,
}}

\title{\method{}: Constructing Multimodal Guidance for Video Professional Motion Editing}

\author[1,2,*]{Siyi Qian}
\author[2,*,\dagger,\ddagger]{Hanqing Yang}
\author[2]{Weihuang Lin}
\author[3,2]{Zihao Yang}
\author[4]{Yichen Lu}
\author[1]{Yuanhong Zheng}
\author[1,\ddagger]{\protect\\ Sheng Li}

\affiliation[1]{Peking University}
\affiliation[2]{Alibaba Token Hub, Alibaba Group}
\affiliation[3]{University of Science and Technology of China}
\affiliation[4]{Zhejiang University}

\contribution[*]{Equal contribution}
\contribution[\dagger]{Corresponding author}
\contribution[\ddagger]{Project lead}
\metadata[Project page]{\url{https://badaleexxx.github.io/MotionCoach-ProjectPage}}

\newcommand{\method}{MotionCoach}
\newcommand{\benchmark}{VPMEBench}
\definecolor{markgreen}{HTML}{257A57}
\definecolor{markred}{HTML}{B34A54}
\definecolor{oursrow}{HTML}{F6F5F8}
\definecolor{oursblue}{HTML}{E8F1FB}
\definecolor{rankfirstblue}{HTML}{D8EAF9}
\definecolor{ranksecondblue}{HTML}{E7F1FA}
\definecolor{rankthirdblue}{HTML}{F3F8FC}
\definecolor{domainfill}{HTML}{F1F5F8}
\definecolor{domainline}{HTML}{496781}
\definecolor{visualfill}{HTML}{EFF7F5}
\definecolor{visualline}{HTML}{2F716C}
\definecolor{temporalfill}{HTML}{F5F1F7}
\definecolor{temporalline}{HTML}{6F587B}
\definecolor{atomicfill}{HTML}{FAF2EF}
\definecolor{atomicline}{HTML}{9B604C}
\newcommand{\rankfirst}[1]{\cellcolor{rankfirstblue}\textbf{#1}}
\newcommand{\ranksecond}[1]{\cellcolor{ranksecondblue}#1}
\newcommand{\rankthird}[1]{\cellcolor{rankthirdblue}#1}
\newcommand{\yesmark}{\textcolor{markgreen}{\ding{51}}}
\newcommand{\nomark}{\textcolor{markred}{\ding{55}}}
\newcommand{\tablechip}[3]{%
  \tikz[baseline=(pill.base)]\node[
    draw=#2,
    fill=#1,
    text=#2,
    rounded corners=1.35pt,
    line width=0.55pt,
    inner xsep=3.8pt,
    inner ysep=0.9pt,
    font=\bfseries\scriptsize
  ] (pill) {#3};%
}
\newcommand{\domainpill}[1]{\tablechip{domainfill}{domainline}{#1}}
\newcommand{\visualpill}[1]{\tablechip{visualfill}{visualline}{#1}}
\newcommand{\temporalpill}[1]{\tablechip{temporalfill}{temporalline}{#1}}
\newcommand{\atomicpill}[1]{\tablechip{atomicfill}{atomicline}{#1}}
\newcommand{\tablevalue}[1]{{\fontsize{8.6}{10.2}\selectfont #1}}
\newcommand{\tableheader}[1]{{\fontsize{8.6}{10.2}\selectfont\bfseries #1}}

\abstract{%
Video editing models achieve strong performance in modifying appearance, backgrounds, styles, and motions, empowered by large-scale training data.
Beyond general scenarios, professional motions are significant in performance, sports, and martial arts. This paper focuses on professional motion editing, which involves fine-grained motion technique and temporal structure.
Given professional motion names but lacking detailed descriptions, existing models often fail to generate professional motions, such as incorrectly lifting the front wheel under a Stoppie instruction.
To address this challenge, we introduce \method{}, a trained multimodal agent that uses multi-turn tools and performs self-verification, producing a grounded prompt and a reference clip as reliable and sufficient multimodal guidance to inject external professional motion knowledge into video editing models.
To supply training data, we build a data pipeline to construct trajectories. We cold-start the agent on \method{}-SFT-10k and apply Group Relative Policy Optimization (GRPO) on \method{}-RL-2k with rewards on the motion guidance prompt, the motion outcome video, and the format.
For evaluation, we introduce \benchmark{}, comprising 314 professional motion editing tasks with fine-grained references and annotations.
Extensive experiments show that \method{} improves the overall score from 55.1 to 65.6 on Seedance 2.0 and from 51.2 to 64.5 on MiniMax-H3.
Learning to construct multimodal guidance enables precise professional motion editing without retraining video editing models.
}

\begin{document}

\thispagestyle{firstheader}
\maketitle
\pagestyle{empty}

\section{Introduction}
\label{sec:introduction}

\begin{figure}[t]
  \centering
  \includegraphics[width=\linewidth]{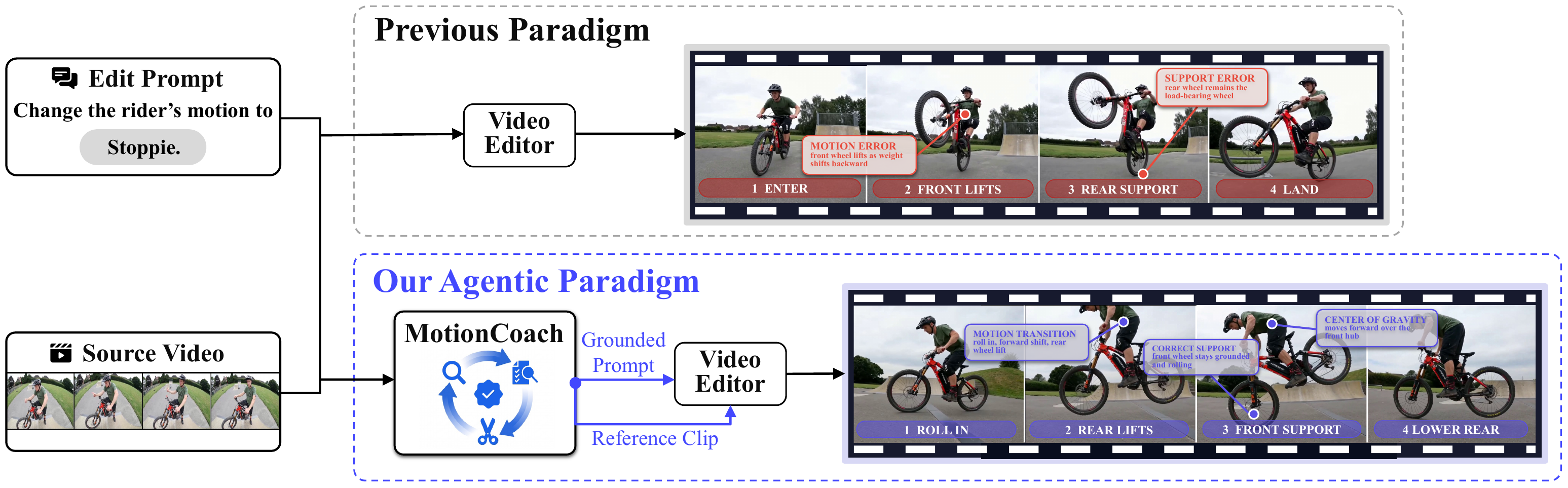}

  \caption{\textbf{Motivation for professional motion editing.} Prompting Seedance 2.0~\citep{bytedanceseed2026seedance2} with \emph{Change the rider's motion to Stoppie} yields a Wheelie (wrong wheel lifted). \method{} instead supplies a grounded prompt and reference clip specifying the correct technique.}
  \label{fig:motivation}
\end{figure}

Existing video editing methods make advances on object replacement, background modification, and style transfer, benefiting from large-scale training data~\citep{cheng2024insv2v,jiang2025vace,bai2025ditto,lin2026kiwiedit,decart2025lucyedit,alibaba2026happyhorse,bytedanceseed2026seedance2,minimax2026h3}.
Beyond general scenarios, we study video professional motion editing, where edited videos must show technically demanding and complex motions while preserving other source content.
Professional motions are typically composed of named atomic motions that can be independently identified, such as Twirl \& Roll in locking and White Crane Spreads Its Wings in tai chi.
However, existing video editing models often generate an incorrect motion.
For example, a Stoppie request instead produces a Wheelie, reversing the required support relationship (Figure~\ref{fig:motivation}).
Correct motions require appropriate technique and temporal structure, including body coordination, support, and stage order or a key pose. A short instruction rarely specifies these details, while the editing models lack the knowledge coverage required for professional motions in vertical scenarios.

Addressing this gap requires supplying external guidance that specifies the professional motion technique. Constructing reliable and sufficient guidance is challenging, which requires presenting the target motion correctly, clearly, and completely.
Some video editing models supplement text instructions with references for concrete visual details~\citep{jiang2025vace,lin2026kiwiedit}, but standard task formulations assume that suitable reference is already available, leaving guidance acquisition and verification outside the task.
Some visual-generation methods use search to acquire external information~\citep{bian2026rsgen}, but relevance alone does not guarantee reliable guidance.
Together, these limitations leave the problem of constructing reliable and sufficient guidance unresolved.

To address this problem, we introduce \textbf{\method{}}, a trained autonomous multimodal agent that learns to perform multi-turn tool use and verify references to obtain reliable and sufficient guidance for professional motion editing.
The agent searches for professional demonstrations, inspects candidates for the target technique, and clips a clear target-motion interval.
Self-verification then assesses motion correctness, visibility, and completeness from the clip, triggering reselection or temporal adjustment.
The agent constructs a grounded prompt from the verified demonstration and source video.
Together, the verified clip and grounded prompt form generator-friendly guidance that conveys the motion technique and temporal structure to the video editing model.

Given the scarcity of multi-turn agent trajectories for professional motion guidance, we build a data pipeline where a teacher model generates trajectories with tools and self-verification, filtered by vision-language model (VLM) checks and human spot-checking, yielding \textbf{\method{}-SFT-10k} and \textbf{\method{}-RL-2k}.
Based on these, we train the agent in two stages.
Supervised fine-tuning (SFT) equips the agent with basic formatting and tool-use capabilities, which are subsequently refined through agentic reinforcement learning (RL) via Group Relative Policy Optimization (GRPO)~\citep{shao2024deepseekmath}.
Specifically, RL leverages three rewards for the motion guidance prompt, the motion outcome video, and the format, with the video reward scoring what the frozen editing model produces in each rollout.
The feedback updates only the agent, improving its selection and organization of reliable and sufficient motion knowledge.

Additionally, to evaluate professional motion correctness in edited videos, we introduce \textbf{\benchmark{}}, 314 human-verified editing tasks across 25 professional motion categories in dance, sports, and martial arts.
We construct the benchmark using professional demonstrations, VLM verification, and human annotations.
Each task provides evaluation references, namely a verified reference clip, selected semantic keyframes, and annotations of temporal structure, motion technique, and negative constraints.
Evaluation uses a VLM as the judge, scoring motion professionalism (MP), physical plausibility (PP), and overall video quality (VQ) against these references.

Experiments show that \method{} improves professional motion editing.
On \benchmark{}, overall scores increase from 55.1 to 65.6 for Seedance 2.0 and from 51.2 to 64.5 for MiniMax-H3 on a 100-point scale.
Ablations support the contribution of agent training and the benefits of combining grounded prompt with reference clip for motion professionalism.
Constructing multimodal guidance with an agent therefore secures guidance reliability and sufficiency, raising the professional correctness of edited motions.

The main contributions are summarized as follows.
\begin{itemize}
\setlength{\itemsep}{0pt}\setlength{\parsep}{0pt}\setlength{\parskip}{0pt}
\item \textbf{\method{}}, an agentic framework that performs multi-turn tool use and self-verification to construct reliable and sufficient multimodal guidance for video editing models.
  \item \textbf{\method{}-SFT-10k}, \textbf{\method{}-RL-2k}, and \textbf{VPMEBench}. We construct datasets for training alongside a human-verified benchmark for professional motion evaluation.
  \item Overall performance gains across MiniMax-H3 and Seedance 2.0 on VPMEBench, empowering video editing models to enhance professional motion editing performance.
\end{itemize}

\section{Related Work}
\label{sec:related_work}

\begin{figure}[t]
  \centering
  \includegraphics[width=\linewidth]{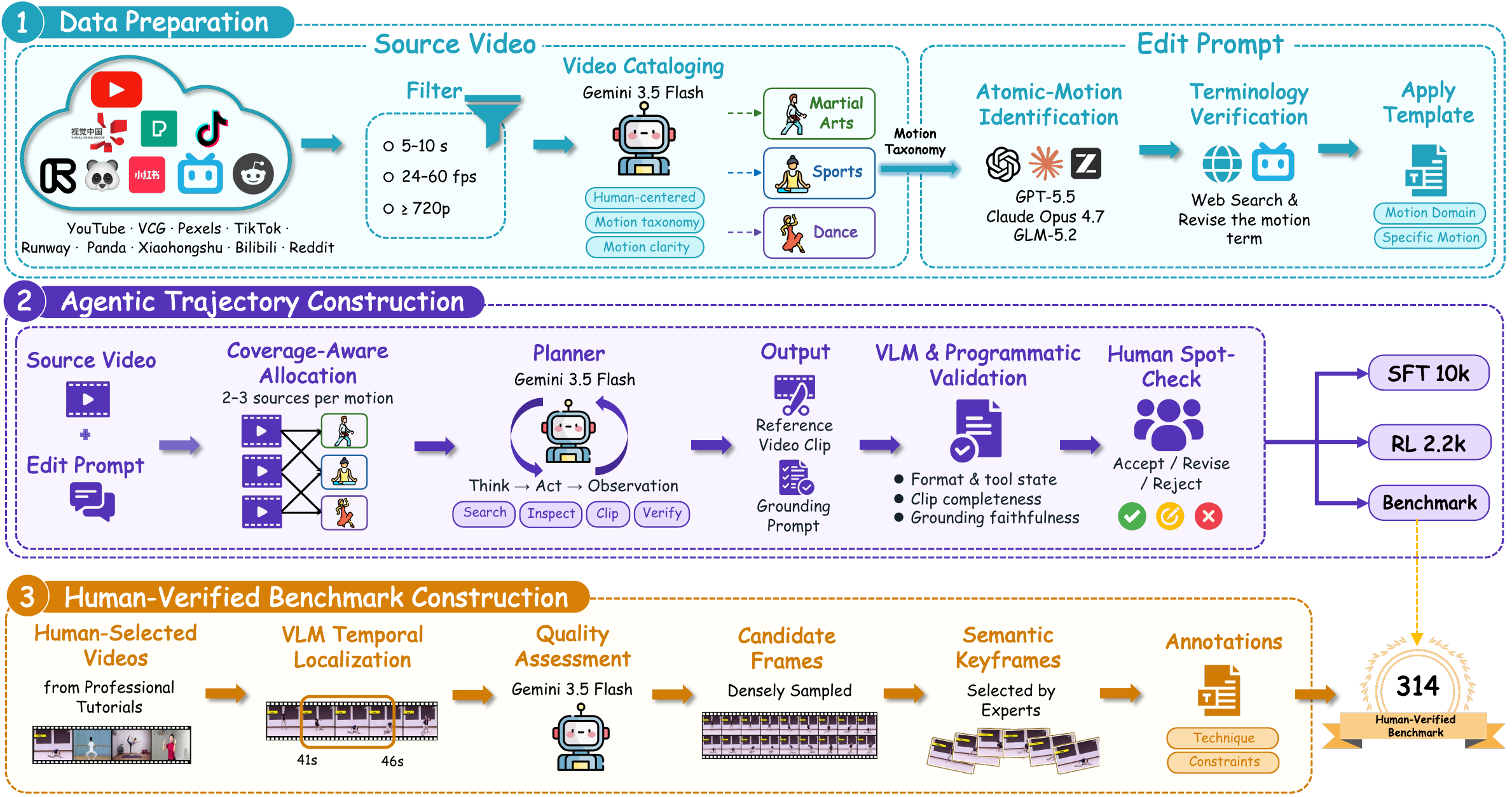}

  \caption{\textbf{Illustration of the data pipeline.} (1) Data preparation prepares source videos and edit prompts. (2) Agentic trajectory construction acquires and verifies motion references to produce \method{}-SFT-10k and \method{}-RL-2k. (3) Human-verified benchmark construction establishes \benchmark{}.}
  \label{fig:data_pipeline}
\end{figure}

\label{sec:related_video_editing}
\textbf{Video editing.} Video editing models support increasingly diverse changes.
Among open-source approaches, InsV2V and Ditto learn instruction-guided editing from synthetic paired data~\citep{cheng2024insv2v,bai2025ditto}, Lucy-Edit-Dev targets clothing, character, and scene changes~\citep{decart2025lucyedit}, and VACE and Kiwi-Edit support reference-conditioned editing~\citep{jiang2025vace,lin2026kiwiedit}.
Richer multimodal inputs also support general-purpose generation and editing in MiniMax-H3 and the closed-source Seedance 2.0 and HappyHorse 1.0~\citep{minimax2026h3,bytedanceseed2026seedance2,alibaba2026happyhorse}. However, professional motion editing remains challenging for these models, which often fail to generate technically demanding motions.

\label{sec:related_agentic_visual}
\label{sec:related_agentic_video}
\textbf{Agentic visual generation and editing.} One line of agentic methods orchestrates pretrained models at inference time, using planning, feedback, and external knowledge to organize visual generation and editing~\citep{yang2026m3,ye2026agentbanana,yao2026photoagent,he2026gems,wu2025movieagent}.
Among these, RS-Gen retrieves external knowledge for image generation, and VideoAgent combines shot planning, asset search, and temporal trimming for video editing~\citep{bian2026rsgen,zhou2026videoagent}.
Another line trains agent policies to connect decisions with visual outcomes~\citep{jiang2026genagent,lin2025jarvisevo,chen2026genevolve,zhang2026weagent}. Agentic methods are increasingly used in image generation, but video generation and editing remain comparatively underexplored.
These results motivate learning to construct editing conditions, while motion technique and temporal structure stay unverified.

\section{Method}
\label{sec:method}

\subsection{Data Construction}
\label{sec:data_construction}

\begin{figure}[t]
  \centering
  \includegraphics[width=\linewidth]{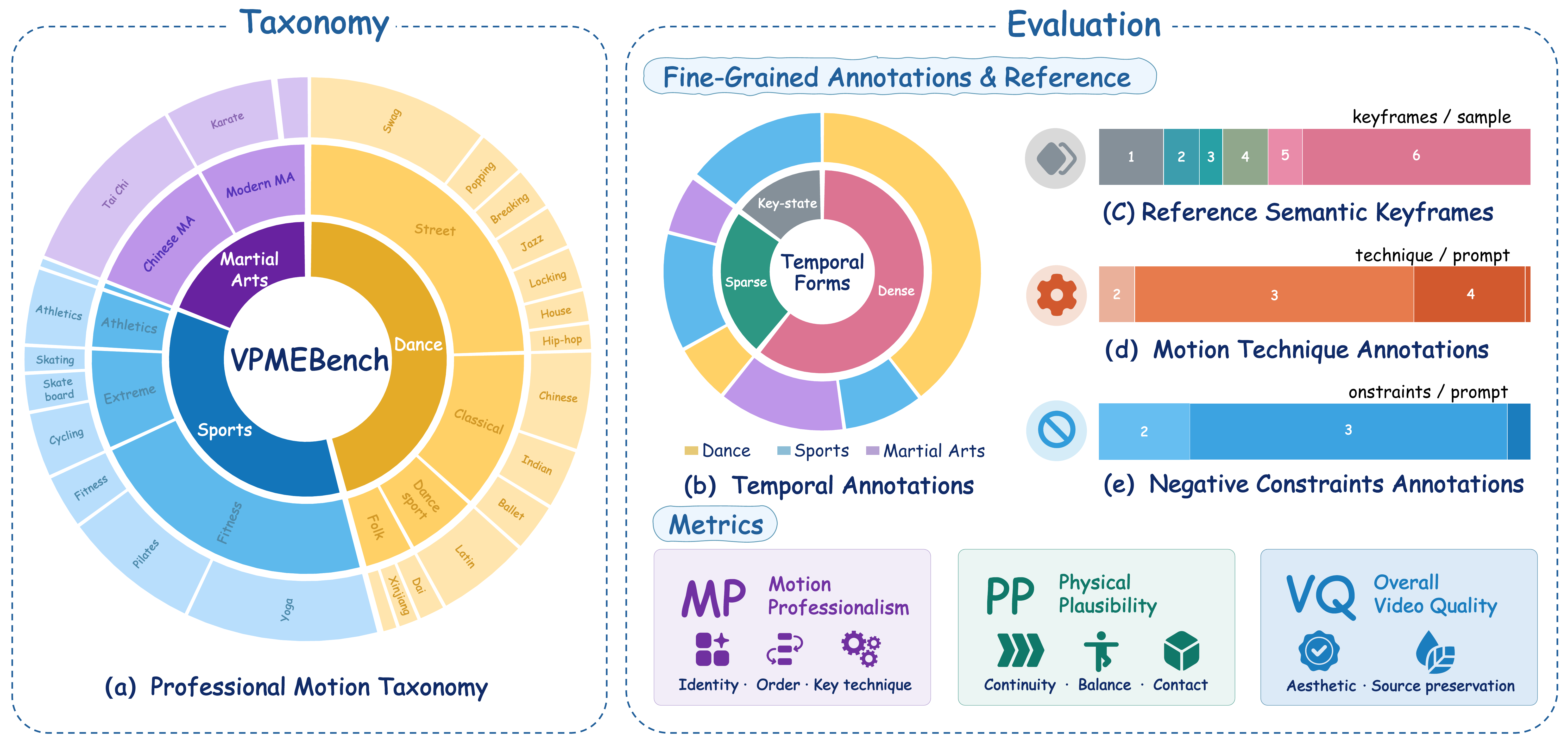}

  \caption{\textbf{Overview of \benchmark{}}. 314 professional motion editing tasks under a hierarchical taxonomy over dance, sports, and martial arts. Each task provides visual references and textual annotations of the target motion for evaluation along motion professionalism (MP), physical plausibility (PP), and overall video quality (VQ).}
  \label{fig:bench}
\end{figure}

To provide video editing models with reliable and sufficient multimodal guidance, a professional motion editing agent must reason over video observations, execute tools, and verify references, but task-specific multi-turn data for learning this workflow remain scarce. Therefore, we design the three-stage pipeline in Figure~\ref{fig:data_pipeline}, which prepares source videos and edit prompts, constructs trajectories that acquire and verify motion guidane, and establishes \benchmark{}.

\textbf{Data preparation and trajectory construction.} We collect videos from online platforms, lasting 5 to 10 seconds at 24 to 60 fps and at 720p or above.
Gemini 3.5 Flash~\citep{kavukcuoglu2026gemini35} then selects human-centered videos with clear motion and classifies them into a motion taxonomy of martial arts, sports, and dance.
For each subcategory, GPT-5.5, Claude Opus 4.7, and GLM-5.2~\citep{openai2026gpt55,anthropic2026claudeopus47,zai2026glm52} propose professional atomic motions and verify the names through web search, and each verified motion fills a short edit-prompt template that defines the editing target of a task. To form complete professional motion editing tasks, we pair source videos with edit prompts, assigning two to three source videos per motion to balance categories.
Gemini 3.5 Flash then generates multi-turn ReAct trajectories~\citep{yao2023react} with search, inspect, and clip tool calls, self-checking the clip to produce the reference clip and the grounded prompt. We use programmatic checks to validate tool-call formats and states. A VLM then assesses clip completeness and grounding faithfulness. Finally, human spot-checks filter the trajectories. The retained trajectories form \method{}-SFT-10k and \method{}-RL-2k. The SFT, RL, and benchmark splits have disjoint target motion names, source videos, and paired edit prompts.

\label{sec:benchmark}

\textbf{Benchmark construction.} General video editing benchmarks do not provide comparable references or annotations for named professional atomic motions (Table~\ref{tab:benchmark_comparison} in Appendix~\ref{app:benchmark}). To enable fine-grained evaluation of professional motion editing, we construct \benchmark{} with references to motion technique and temporal structure. Human annotators select demonstrations from professional tutorials on online platforms for each subcategory. A VLM locates suitable target-motion intervals, and experts select up to six semantic keyframes from densely sampled frames per clip.
\benchmark{} contains 314 professional motion editing tasks across 3 top-level, 10 second-level, and 25 leaf categories, each taking a source video and an edit prompt as input and providing fine-grained visual references and textual annotations of the target motion (Figure~\ref{fig:bench} and Appendix~\ref{app:benchmark}).

We use a VLM as the judge, scoring each dimension 0, 0.5, or 1 from the generated video, source video, and task references (Appendix~\ref{app:benchmark}).
\textbf{Motion professionalism (MP)} assesses motion technique and temporal structure, including stage order or a key pose, while accepting equivalent executions.
\textbf{Physical plausibility (PP)} assesses motion continuity, balance, limb integrity, and contact consistency.
\textbf{Overall video quality (VQ)} combines video quality and edit preservation at a ratio of 1:2, covering clarity, temporal stability, and aesthetics as well as preservation of the subject, scene, and non-target content beyond the required motion change.

The overall score is
\begin{equation}
\mathrm{Overall}=w_{\mathrm{MP}}\,\mathrm{MP}+w_{\mathrm{PP}}\,\mathrm{PP}+w_{\mathrm{VQ}}\,\mathrm{VQ},
\label{eq:benchmark_score}
\end{equation}
To emphasize the importance of MP for this task, we set the weights to $w_{\mathrm{MP}}=0.5$, $w_{\mathrm{PP}}=0.2$, and $w_{\mathrm{VQ}}=0.3$.

\begin{figure}[t]
  \centering
  \includegraphics[width=\linewidth]{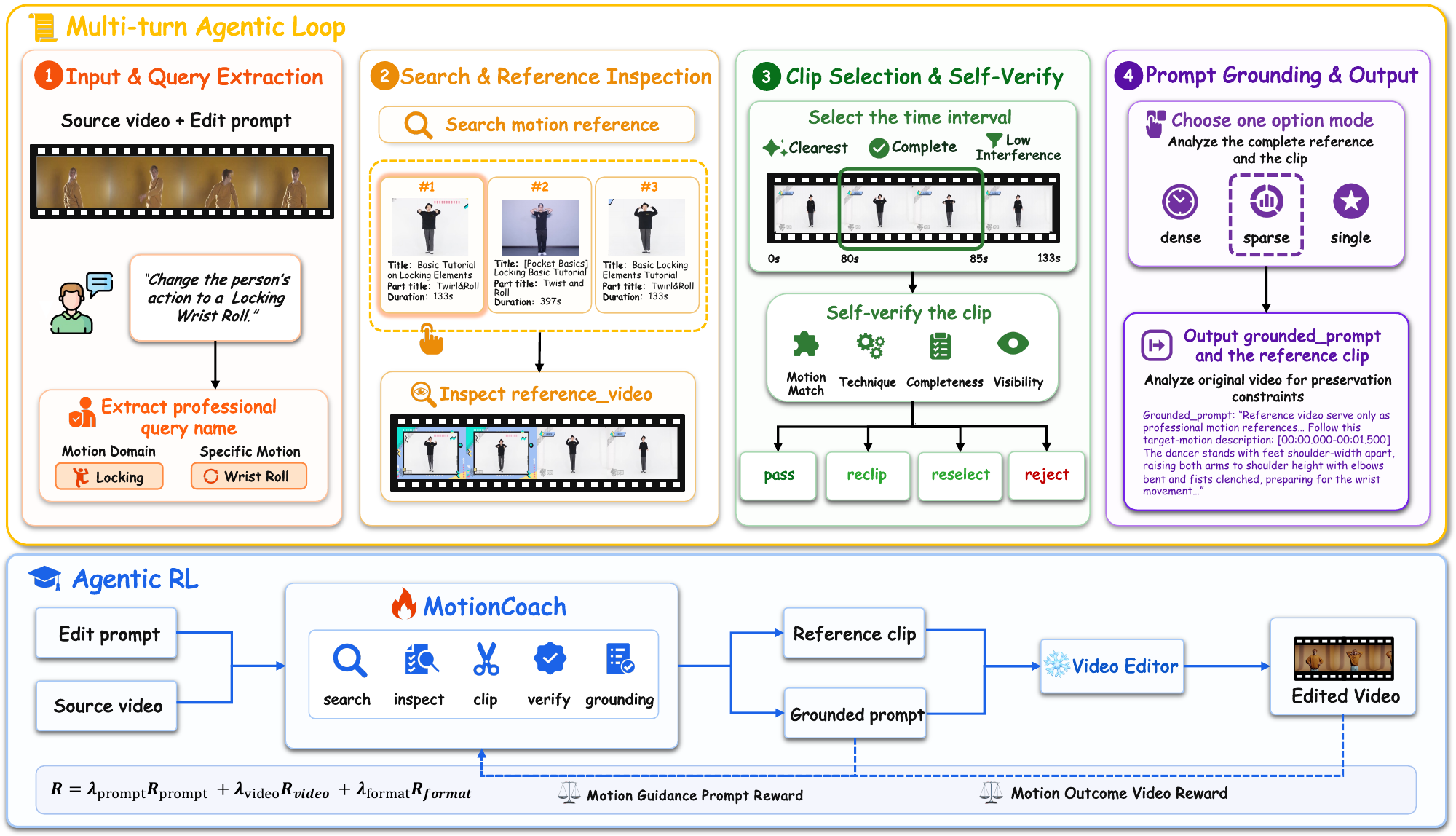}

  \caption{\textbf{Overview of \method{}}. Top: the multi-turn agentic loop during rollout and inference uses search, inspect, and clip tools with self-verification to produce reliable and sufficient motion guidance. Bottom: Agentic RL updates the policy with motion guidance prompt, motion outcome video, and format rewards while keeping the video editing model frozen.}
  \label{fig:method}
\end{figure}

\subsection{Agentic Framework}
\label{sec:agentic_framework}
\label{sec:task}

Professional motion editing requires motion technique and temporal structure.
To provide richer external motion knowledge, \method{} takes a source video $V_s$ and an edit prompt $u$, calls the search, inspect, and clip tools, and verifies the references itself to produce a grounded prompt $p$ and a verified reference clip $C_r$ as multimodal guidance (Figure~\ref{fig:method} and Appendix~\ref{app:system_prompt}). A frozen video editing model $G$ then generates $\hat V=G(V_s,p,C_r)$.

\textbf{Search and reference inspection.} Since training data for professional motion editing is scarce, generative models lack the necessary prior knowledge. To provide external knowledge to compensate for this, the agent actively collect external references. Specifically, the agent extracts the motion domain and specific name from the edit prompt and calls \textbf{search} to obtain candidates from video platforms. The agent then uses \textbf{inspect} to confirm the motion technique of the complete reference video $V_r$ from visible demonstration, explanations, subtitles, and on-screen text.

\textbf{Clip selection and self-verification.} Complete reference videos are often noisy and too long for most video editing models to handle. To obtain the core motion technique, the agent selects a clear, complete, and low-interference interval from the complete reference video $V_r$ that shows the target motion and obtains a verified reference clip $C_r$:
\begin{equation}
C_r=\operatorname{Clip}(V_r,[t_s,t_e]),\qquad C_r\in\mathcal{C}_{\mathrm{valid}}.
\label{eq:verified_clip}
\end{equation}
Here, $\mathcal{C}_{\mathrm{valid}}$ denotes clips accepted by self-verification. If $C_r\notin\mathcal{C}_{\mathrm{valid}}$, the agent re-clips $V_r$ or reselects another candidate to ensure reference reliability.

\textbf{Grounded prompt construction.} The agent uses the complete reference video $V_r$ and verified reference clip $C_r$ to obtain sufficient professional motion knowledge.
The agent then selects one of three description formats: dense temporal, sparse stage, or key-pose.

Let $D(V_r,C_r)$ denotes the motion description, covering motion technique, temporal structure, and motion-specific errors to avoid, and let $P(V_s)$ require all other source content to remain unchanged.
The final answer outputs a verified reference clip $C_r$ and a grounded prompt $p$. The grounded prompt is
\begin{equation}
p=\operatorname{Compose}\bigl(u,D(V_r,C_r),P(V_s)\bigr),
\label{eq:grounded_prompt}
\end{equation}
where $\operatorname{Compose}$ assembles these elements into one natural language prompt.

\subsection{Training Strategy}
\label{sec:training}

Our agent policy learns to acquire reliable, sufficient, and generator-friendly motion guidance in two stages.
We cold-start the policy on \method{}-SFT-10k to establish multi-turn formatting and basic tool use, and agentic RL on \method{}-RL-2k then improves the policy through the quality of the resulting guidance and the frozen model's editing performance.

During the RL stage, to check whether the constructed motion guidance is precise and sufficient, we apply a motion guidance prompt reward $R_{\mathrm{prompt}}$ that compares the grounded prompt $p$ from the agent policy with a teacher prompt $p^\star$,
\begin{equation}
R_{\mathrm{prompt}}=J_{\mathrm{prompt}}(V_s,u,p^\star,p).
\label{eq:prompt_reward}
\end{equation}
The function $J_{\mathrm{prompt}}$ scores motion correctness and specificity, structural completeness, edit scope, and clarity, then aggregates the scores with dimension-specific weights (Appendix~\ref{app:prompt_reward_judge}).
The teacher prompt guides structure and content coverage while equivalent wording remains acceptable.

To judge whether the constructed multimodal guidance actually supports correct professional motion editing, we apply a motion outcome video reward $R_{\mathrm{video}}$, where $J_{\mathrm{video}}$ scores the outcome $\hat V$ that the frozen editing model produces from $V_s$, $p$, and $C_r$ against a teacher reference $R$ that represents motion knowledge,
\begin{equation}
R_{\mathrm{video}}=J_{\mathrm{video}}(V_s,u,R,\hat V).
\label{eq:video_reward}
\end{equation}
Motion professionalism, physical plausibility, visual quality, and edit preservation each receive a three-level score, and the weighted aggregate emphasizes professional performance while also considering physical plausibility and edit preservation (Appendix~\ref{app:video_reward_judge}).

To secure valid tool interaction and structured outputs as a basic capability, we apply a format reward $R_{\mathrm{format}}$ that equals 1 when every tool call uses a valid tool name and JSON arguments and the final answer contains a nonempty grounded prompt and one reference-clip identifier, and 0 otherwise.
The three components lie in $[0,1]$ and combine as
\begin{equation}
R=\lambda_{\mathrm{prompt}}R_{\mathrm{prompt}}+\lambda_{\mathrm{video}}R_{\mathrm{video}}+\lambda_{\mathrm{format}}R_{\mathrm{format}},
\label{eq:reward}
\end{equation}
where the format term is added separately and the weight setting follows the ablation in Table~\ref{tab:reward_weights}. We optimize the agent policy on this reward with GRPO. Samples affected by generation or judging infrastructure failures are excluded to avoid attributing service errors to the agent policy.

\section{Experiments}
\label{sec:experiments}

\subsection{Experimental Setup}

\textbf{Training details.} We train \method{} with Qwen3-VL-8B-Instruct~\citep{bai2025qwen3vl}, first cold-starting the policy with SFT on \method{}-SFT-10k (AdamW, learning rate $10^{-5}$, all parameters, 2 epochs).
For RL, we run GRPO on \method{}-RL-2k (AdamW, learning rate $10^{-6}$, task mini-batch size of 8) with 16 GPUs, plus 8 more GPUs serving INT8 MiniMax-H3 for rollout video generation. Qwen3-VL-30B-A3B serves as the judge for RL rewards.
RL takes approximately 28 hours.
Full training details are in Appendix~\ref{app:training}.

\textbf{Benchmark and evaluation.} We evaluate \method{} on \benchmark{}, which measures professional motion editing capability, with 314 samples covering dance, sports, and martial arts.
We compare open-source Ditto~\citep{bai2025ditto}, Kiwi-Edit~\citep{lin2026kiwiedit}, InsV2V~\citep{cheng2024insv2v}, Lucy-Edit-Dev~\citep{decart2025lucyedit}, MiniMax-H3~\citep{minimax2026h3} and closed-source Seedance 2.0~\citep{bytedanceseed2026seedance2}, HappyHorse 1.0~\citep{alibaba2026happyhorse}, applying \method{} on the frozen MiniMax-H3 and Seedance 2.0.
Following Section~\ref{sec:benchmark}, we use a multimodal judge accessed through an API to score generated videos.
We report motion professionalism (MP), physical plausibility (PP), overall video quality (VQ), and the overall score on a 100-point scale.
The overall score follows Equation~\ref{eq:benchmark_score}. Human annotators provide pairwise judgments. QWK reaches 0.846 overall and above 0.73 on every dimension, supporting agreement between the automatic evaluation and human judgments (Table~\ref{tab:judge_human_agreement} and Appendix~\ref{app:benchmark}).

\begin{table*}[t]
  \caption{\textbf{Professional motion editing on \benchmark{}}. Existing video editing models use short edit prompts. We evaluate \method{} on frozen MiniMax-H3 and Seedance 2.0, where the agent provides multimodal guidance for each model.}
  \label{tab:main}
  \centering
  \small
  \setlength{\tabcolsep}{3.2pt}
  \begin{tabular}{lccccccccccccc}
    \toprule
    \textbf{Model} & \multicolumn{3}{c}{Dance} & \multicolumn{3}{c}{Sports} & \multicolumn{3}{c}{Martial Arts} & \multicolumn{3}{c}{All} & \textbf{Overall} \\
    \cmidrule(lr){2-4} \cmidrule(lr){5-7} \cmidrule(lr){8-10} \cmidrule(lr){11-13}
     & MP & PP & VQ & MP & PP & VQ & MP & PP & VQ & MP & PP & VQ &  \\
    \midrule
    \multicolumn{14}{l}{\textbf{\textit{Open-source models}}} \\
    Ditto & 0.7 & \rankthird{88.2} & 59.5 & 0.9 & \rankfirst{94.1} & 79.1 & 0.0 & \ranksecond{88.3} & 63.9 & 0.6 & \ranksecond{90.3} & 67.2 & 38.5 \\
    Kiwi-Edit & 1.7 & 61.8 & 72.6 & 0.9 & 81.4 & \rankthird{91.2} & 0.0 & 72.5 & 79.7 & 1.1 & 70.7 & 80.5 & 38.8 \\
    InsV2V & 0.3 & 82.6 & 78.1 & 0.0 & 87.7 & 88.9 & 0.0 & \rankthird{85.8} & 85.8 & 0.2 & 85.0 & 83.4 & 42.1 \\
    Lucy-Edit-Dev & 0.0 & 85.8 & 86.3 & 0.0 & \rankthird{88.2} & 87.0 & 0.0 & 85.0 & 84.2 & 0.0 & 86.5 & 86.1 & 43.1 \\
    MiniMax-H3 & 12.8 & \ranksecond{88.5} & \rankthird{88.9} & 24.1 & 80.0 & 85.8 & 10.0 & 84.2 & \rankthird{86.1} & 16.2 & 84.7 & \rankthird{87.3} & 51.2 \\
    \textbf{MiniMax-H3 + Ours} & \ranksecond{29.2} & \rankfirst{91.7} & \rankfirst{99.3} & \rankthird{44.5} & \ranksecond{92.7} & \rankfirst{98.0} & \rankthird{21.7} & \rankfirst{92.5} & \rankfirst{95.8} & \rankthird{33.1} & \rankfirst{92.2} & \rankfirst{98.2} & \ranksecond{64.5} \\
    \midrule
    \multicolumn{14}{l}{\textbf{\textit{Closed-source models}}} \\
    HappyHorse 1.0 & \rankthird{26.4} & 73.6 & \ranksecond{90.2} & 40.9 & 78.2 & \ranksecond{93.5} & \ranksecond{45.8} & 78.3 & \ranksecond{88.1} & \ranksecond{35.2} & 76.1 & \ranksecond{90.9} & \rankthird{60.1} \\
    Seedance 2.0 & 18.8 & 85.8 & 70.0 & \ranksecond{48.6} & 87.3 & 90.5 & 20.8 & 83.3 & 69.7 & 29.6 & 85.8 & 77.1 & 55.1 \\
    \textbf{Seedance 2.0 + Ours} & \rankfirst{46.2} & 85.4 & 71.1 & \rankfirst{61.4} & 86.4 & 73.8 & \rankfirst{54.2} & \rankfirst{92.5} & 72.2 & \rankfirst{53.0} & \rankthird{87.1} & 72.2 & \rankfirst{65.6} \\
    \bottomrule
  \end{tabular}
\end{table*}

\begin{table}[t]
  \begin{minipage}[t]{0.30\linewidth}
    \centering
    \footnotesize
    \caption{Ablation of workflow and training stages.}
    \label{tab:training_stages}
    \setlength{\tabcolsep}{2pt}
    \begin{tabular}{lcc}
      \toprule
      \textbf{Variant} & \textbf{SD2} & \textbf{H3} \\
      \midrule
      Base            & 55.1 & 51.2 \\
      Base + Workflow & 54.0 & 52.2 \\
      Base + SFT      & 62.3 & 62.8 \\
      Base + RL       & \textbf{65.6} & \textbf{64.5} \\
      \bottomrule
    \end{tabular}
  \end{minipage}%
  \hfill
  \begin{minipage}[t]{0.36\linewidth}
    \centering
    \footnotesize
    \caption{Multimodal input ablation for frozen video editing models.}
    \label{tab:input_ablation}
    \setlength{\tabcolsep}{2pt}
    \begin{tabular}{ccccc}
      \toprule
      \textbf{Var.} & \textbf{Ref. Clip} & \textbf{Prompt} & \textbf{SD2} & \textbf{H3} \\
      \midrule
      A & \nomark & \yesmark & 59.8 & 63.7 \\
      B & \yesmark & \nomark & 52.4 & 48.6 \\
      \rowcolor{oursblue}
      C & \yesmark & \yesmark & \textbf{65.6} & \textbf{64.5} \\
      D & \yesmark$^\dagger$ & \yesmark$^\dagger$ & 72.1 & 68.9 \\
      \bottomrule
    \end{tabular}
  \end{minipage}%
  \hfill
    \begin{minipage}[t]{0.28\linewidth}
    \centering
    \footnotesize
    \caption{Ablation of agent components.}
    \label{tab:agent_component_ablation}
    \setlength{\tabcolsep}{2pt}
    \begin{tabular}{lcc}
      \toprule
      \textbf{Variant} & \textbf{SD2} & \textbf{H3} \\
      \midrule
      Ours & \textbf{65.6} & \textbf{64.5} \\
      w/o Video & 54.9 & 51.3 \\
      w/o Clip & 57.4 & 54.1 \\
      w/o Verify & 58.1 & 54.6 \\
      \bottomrule
    \end{tabular}
  \end{minipage}

  \par\vspace{4pt}

  \caption{Policy and tool execution, in percent.}
  \label{tab:agent_stage_accuracy}
  \centering
  \small
  \setlength{\tabcolsep}{5pt}
  \begin{tabular*}{\linewidth}{@{\extracolsep{\fill}}lccccccc@{}}
    \toprule
    & \multirow{2}{*}{\textbf{Query Extr.}} & \multicolumn{5}{c}{\textbf{Tool-Call Success}} & \multirow{2}{*}{\textbf{Traj.}} \\
    \cmidrule(lr){3-7}
    & & All & Search & Inspect & Clip & Pass@1 & \\
    \midrule
    Base & 80.9 & 82.5 & 99.5 & 99.5 & 62.2 & 77.4 & 90.1 \\
    Ours & \textbf{93.6} & \textbf{99.4} & \textbf{100.0} & \textbf{100.0} & \textbf{98.3} & \textbf{95.9} & \textbf{99.7} \\
    \bottomrule
  \end{tabular*}

\end{table}

\begin{figure}[!t]
  \centering
  \includegraphics[width=0.72\linewidth]{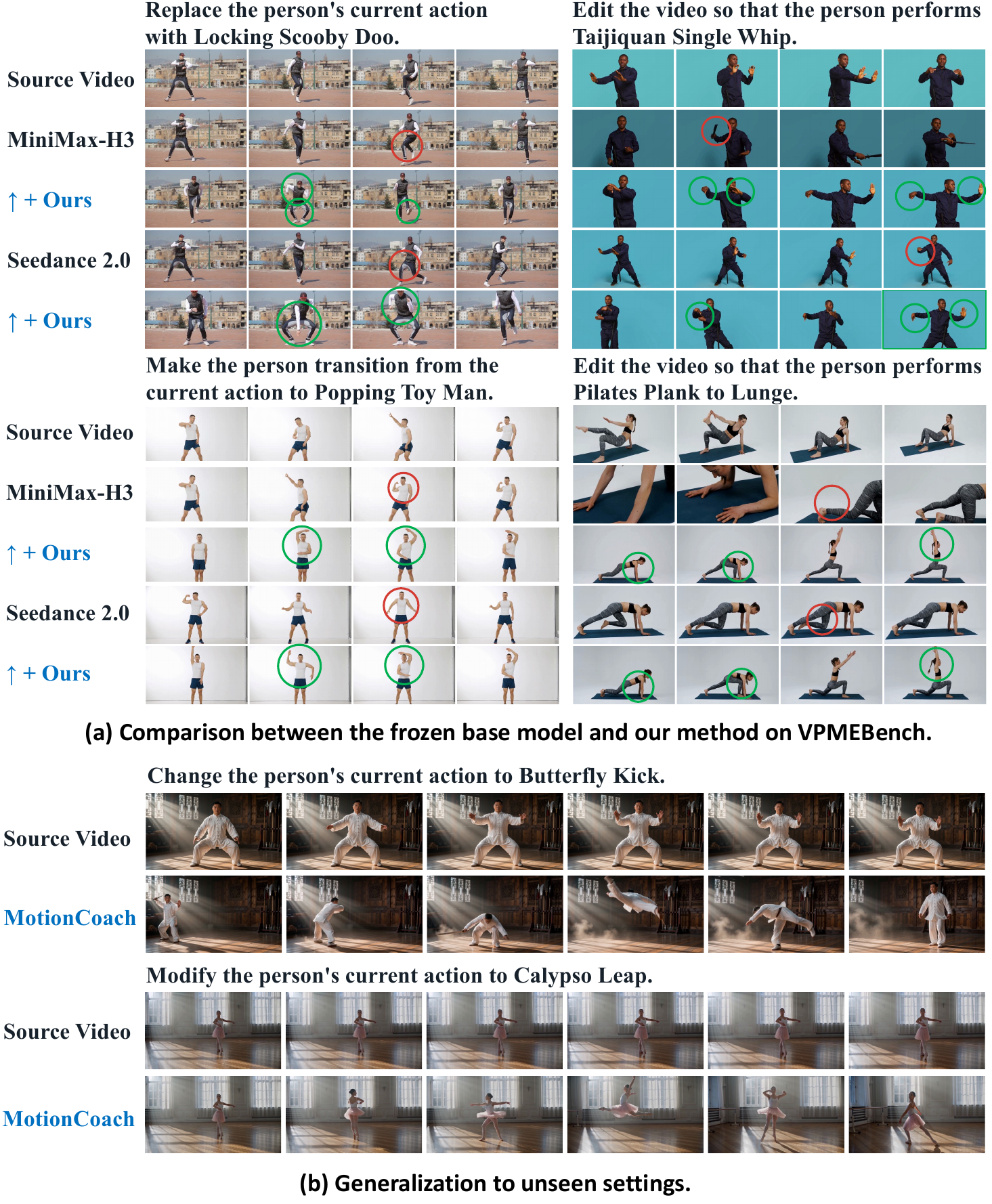}

  \caption{Qualitative results of \method{}.}
  \label{fig:qualitative}
\end{figure}

\subsection{Main Results on \benchmark{}}

\textbf{Quantitative results.} As shown in Table~\ref{tab:main}, professional motion editing challenges both open-source and closed-source video editing models: all open-source models except MiniMax-H3 score below 50 overall, while Seedance 2.0 and HappyHorse 1.0 reach only 55.1 and 60.1 overall. Complex atomic motions often use names rare in training data, so existing models may retain the source motion, leading to low MP.
Applying \method{} substantially improves the frozen MiniMax-H3 across all reported metrics. Furthermore, when transferred to Seedance~2.0, our method generalizes well across most dimensions, underscoring its plug-and-play applicability. Overall, these results show that the learned guidance-construction strategy generalizes across video editing models.

\textbf{Qualitative results.} As shown in Figure~\ref{fig:qualitative}(a), existing video editing models often retain the source motion unchanged when facing unfamiliar professional motion names. In addition, they may misinterpret the names as unrelated visual concepts, such as generating an actual whip object rather than the Tai Chi motion \textit{Single Whip}. In contrast, equipped with \method{}, video editing models successfully generate accurate body configurations along with proper temporal structures across dance, martial arts, and sports.
Additional qualitative results are provided in Figure~\ref{fig:qualitative_extra} of Appendix~\ref{app:results}.

\subsection{Experimental Analysis}

\textbf{Training strategy.} Table~\ref{tab:training_stages} separates the effects of the workflow and training. The untrained workflow yields inconsistent gains across Seedance~2.0 (SD2) and MiniMax-H3 (H3), showing that tool access alone does not provide reliable improvements. Subsequently, SFT consistently improves performance across both base editors, and RL further refines the SFT policy. The overall gains therefore stem from learning to construct useful guidance rather than merely executing the workflow.

\textbf{Multimodal input conditions.} Table~\ref{tab:input_ablation} shows that removing 
the grounded prompt (B) costs 13.2/15.9 points on Seedance(SD2)/MiniMax-H3(H3), versus 5.8/0.8 for removing the reference clip (A), since the prompt carries motion technique and temporal structure while 
the clip offers complementary visual reference. Notably, MotionCoach closely approaches expert-guided performance, trailing the human-verified guidance (Variant~D) by only 6.5/4.4 points.

\textbf{Agent component.} Table~\ref{tab:agent_component_ablation} isolates the main components. w/o Video, which replaces the reference video with one image, reduces Overall by 10.7 and 13.2 points. w/o Clip, which uniformly samples a five-second segment, causes drops of 8.2 and 10.4 points. w/o Verify, which removes the verify–reclip–reselect step, causes drops of 7.5 and 9.9 points. These results show that temporal video reference, focused clip selection, and explicit self-verification make complementary contributions to reliable guidance construction.

\textbf{Policy and tool execution.} Table~\ref{tab:agent_stage_accuracy} examines where training improves the agent workflow, which reports query accuracy (whether the target-motion name is correctly extracted), tool-call success (whether search, inspect, and clip are executed correctly), Pass@1 (success without retries), and trajectory success. The largest tool-level improvement occurs in clip execution, which increases from 62.2\% to 98.3\%.
Training improves most stages, raising trajectory success to 99.7\%, which supports 
the construction of reliable and sufficient guidance.

\begin{table*}[t]
  \caption{Generalization on the subject-motion editing subset of IVEBench~\citep{chen2026ivebench}.}
  \label{tab:ivebench_generalization}
  \centering
  \scriptsize
  \setlength{\tabcolsep}{1.6pt}
  \begin{tabular*}{\textwidth}{@{\extracolsep{\fill}}lccccccccccc@{}}
    \toprule
    & \multirow{2}{*}{\textbf{Total}} & \multicolumn{10}{c}{\textbf{Metric Performance}} \\
    \cmidrule(lr){3-12}
    \textbf{Model} & & \textbf{SC} & \textbf{BC} & \textbf{TF} & \textbf{MS} & \textbf{VTSS} & \textbf{OSC} & \textbf{PSC} & \textbf{IS} & \textbf{SF} & \textbf{CF} \\
    \midrule
    MiniMax-H3 & 48.0 & 90.9 & \textbf{95.3} & 96.5 & 98.9 & \textbf{6.4} & 19.3 & 19.1 & \textbf{2.73} & 69.5 & 1.40 \\
    \textbf{MiniMax-H3 + Ours} & \textbf{53.8} & \textbf{92.3} & 95.2 & \textbf{97.5} & \textbf{99.0} & 6.1 & \textbf{24.5} & \textbf{19.2} & 2.62 & \textbf{95.3} & \textbf{2.05} \\
    \bottomrule
  \end{tabular*}
\end{table*}

\begin{table}[t]
  \begin{minipage}[t]{0.52\linewidth}
    \centering
    \small
    \caption{Reward weight ablation.}
    \renewcommand{\arraystretch}{1.1}
    \label{tab:reward_weights}
    \setlength{\tabcolsep}{4pt}
    \begin{tabular}{ccccc}
      \toprule
      \shortstack{Prompt:Video} & MP & PP & VQ & \textbf{Overall} \\
      \midrule
      0.0 : 1.0 & 50.47 & 83.43 & 70.12 & 62.96 \\
      0.3 : 0.7 & 52.22 & 85.51 & 69.05 & 63.93 \\
      0.5 : 0.5 & 54.93 & 84.24 & 68.84 & 64.97 \\
      0.7 : 0.3 & 53.03 & 87.10 & 72.24 & \textbf{65.61} \\
      1.0 : 0.0 & 54.46 & 82.96 & 70.22 & 64.89 \\
      \bottomrule
    \end{tabular}
  \end{minipage}%
  \hfill
  \begin{minipage}[t]{0.46\linewidth}
    \centering
    \small
    \caption{Policy comparison with and without task-specific training.}
    \label{tab:policy_scale}
    \setlength{\tabcolsep}{4pt}
    \begin{tabular}{lcc}
      \toprule
      \multirow{2}{*}{\textbf{Policy}} & \multicolumn{2}{c}{\textbf{Overall}} \\
      \cmidrule(lr){2-3}
      & Seedance 2.0 & MiniMax-H3 \\
      \midrule
      Qwen3.5-9B  & 50.00 & 48.87 \\
      Qwen3.5-27B & 57.45 & 56.07 \\
      Ours        & \textbf{65.61} & \textbf{64.46} \\
      \bottomrule
    \end{tabular}
  \end{minipage}

  \caption{Agreement between the automatic evaluation and human judgments.}
  \label{tab:judge_human_agreement}
  \centering
  \small
  \setlength{\tabcolsep}{4pt}
  \begin{tabular}{lcccccc@{\hskip 1.2em}cc}
    \toprule
    & \multicolumn{2}{c}{MP} & \multicolumn{2}{c}{PP} & \multicolumn{2}{c}{VQ} & \multicolumn{2}{c}{Overall} \\
    \cmidrule(lr){2-3} \cmidrule(lr){4-5} \cmidrule(lr){6-7} \cmidrule(lr){8-9}
    & QWK & 95\% CI & QWK & 95\% CI & QWK & 95\% CI & QWK & 95\% CI \\
    \midrule
    Pooled ($n=100$) & 0.803 & [0.71, 0.88] & 0.733 & [0.60, 0.84] & 0.943 & [0.90, 0.97] & 0.846 & [0.74, 0.92] \\
    \bottomrule
  \end{tabular}
\end{table}

\textbf{Reward weight ablation.} Table~\ref{tab:reward_weights} studies the balance between the prompt and video rewards. The 0.7:0.3 ratio achieves the best Overall score of 65.61 and is used in our final model. The prompt reward encourages precise motion specification, while the video reward evaluates whether the resulting guidance can be executed by the frozen editor.

\textbf{Task-specific training versus model scaling.} In Table~\ref{tab:policy_scale}, all policies use the same information, tools, and inference budget. Scaling the untrained policy from 9B to 27B improves Overall from 50.00 to 57.45 on Seedance 2.0 and from 48.87 to 56.07 on MiniMax-H3. Our trained 8B policy reaches 65.61 and 64.46, outperforming the untrained 27B policy by 8.16 and 8.39 points. A larger backbone provides a stronger starting point, but does not replace task-specific training.

\textbf{Generalization to unseen settings.} In Table~\ref{tab:ivebench_generalization}, we evaluate \method{} on the unseen subject-motion editing subset of IVEBench. With frozen MiniMax-H3, \method{} raises Total from 48.0 to 53.8. Additionally, as shown in Figure~\ref{fig:qualitative}(b), we evaluate editing performance on unseen professional motion terms that are absent from both the training data and VPMEBench, where \method{} autonomously gathers domain knowledge to guide the video editing model in executing faithful motion modifications.

\FloatBarrier

\section{Conclusion}
\label{sec:conclusion}

This work focuses on professional motion editing, which requires edited videos to exhibit motion technique and temporal structure. We propose \method{}, an agentic framework that uses multi-turn tools and performs self-verification to get a reference clip and grounded prompt as reliable and sufficient multimodal guidance for a frozen video editing model. We build a trajectory data pipeline and \benchmark{}, and train the policy with agentic RL. Extensive experiments show that \method{} improves professional motion editing without retraining video editing models.

\newpage
\bibliographystyle{plainnat}
\bibliography{references}

\appendix
\newpage
\section{Appendix}

\subsection{Benchmark}
\label{app:benchmark}

\benchmark{} contains 314 professional motion editing tasks across 3 top-level, 10 second-level, and 25 leaf categories.
Over the top-level categories, dance holds 144 tasks (45.9\%), sports and fitness holds 110 tasks (35.0\%), and martial arts holds 60 tasks (19.1\%).

\begin{figure}[p]
  \centering
  \includegraphics[width=\linewidth]{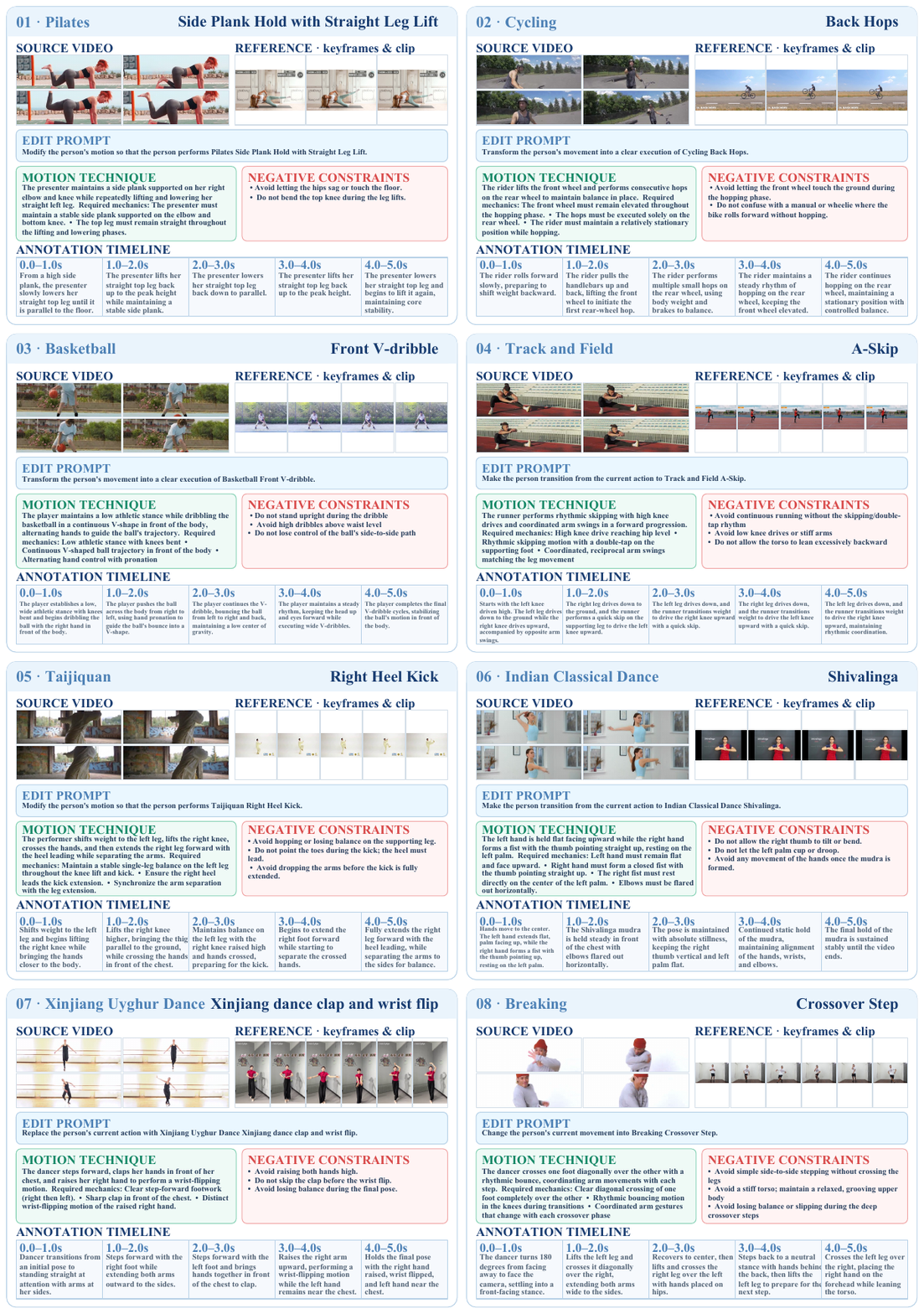}
  \caption{Representative \benchmark{} tasks.}
  \label{fig:bench_cases_page1}
\end{figure}

Each task provides a human-verified five-second reference clip together with the fine-grained annotations summarized in Figure 3(b)–(e). (b) Temporal annotations describe each motion using one of three structure-dependent forms: dense, sparse, or key-state, covering 191, 76, and 47 tasks, respectively. A fast Breaking move such as Noreaga Drop receives a dense per-second timeline, a multi-stage bicycle trick such as Back Hops is divided into a few timestamped sparse stages, and a held yoga pose such as Warrior Two is represented by a single key state without timestamps. (c) Reference semantic keyframes mark one to six defining body configurations, ranging from six selected configurations for a Tai Chi form to a single frame for a static hold, giving 1,403 keyframes across the benchmark. (d) Motion technique annotations list the two to five execution criteria that a correct performance must satisfy, such as pivoting the supporting foot outward during a roundhouse kick. (e) Negative constraint annotations list the two to four common errors to avoid, such as allowing the knee to collapse inward. Table~\ref{tab:benchmark_comparison} compares \benchmark{} with representative video-editing benchmarks. Figures~\ref{fig:bench_cases_page1} show representative \benchmark{} tasks across the dance, sports and fitness, and martial arts top-level categories, together with the source video, edit prompt, reference clip, semantic keyframes, and motion annotations provided for each task.

\textbf{Human validation.} Before annotation, we drew a stratified random sample of 100 paired outputs, with 50 pairs from each editor. Each pair contained Base and Base+RL outputs generated from the same source video and instruction, where Base denotes the frozen editor without MotionCoach and Base+RL uses guidance from the fully trained agent. The sample contained 46 dance, 34 sports, and 20 martial-arts pairs. We recruited eight annotators, each with at least three years of relevant training, coaching, performance, or competitive experience. Each pair was independently evaluated by three annotators assigned according to their domain expertise. Annotators had no role in method or benchmark development. After calibration on held-out examples, they evaluated anonymized outputs presented in randomized A/B order and selected Base preferred, tie, or Base+RL preferred for MP, PP, VQ, and Overall. We aggregated the three judgments for each dimension using their ordinal median (Base preferred $<$ tie $<$ Base+RL preferred). Automatic preferences were derived from $\Delta$, the Base+RL score minus the Base score: $|\Delta|\leq\tau$ was treated as a tie, and the sign of $\Delta$ determined the preference otherwise. We set $\tau=3$ points on the 0--100 scale, based on the evaluator's standard deviation over five repeated judgments on a held-out calibration set. Agreement between automatic and aggregated human preferences was measured using quadratic-weighted kappa, reaching 0.803, 0.733, 0.943, and 0.846 for MP, PP, VQ, and Overall, respectively. The corresponding 95\% confidence intervals are reported in Table~\ref{tab:judge_human_agreement} and were computed using 10,000 bootstrap resamples clustered by benchmark task.

\subsection{Additional Results}

\label{app:results}

\begin{figure}[p]
  \centering
  \includegraphics[width=\linewidth,height=0.455\textheight,keepaspectratio]{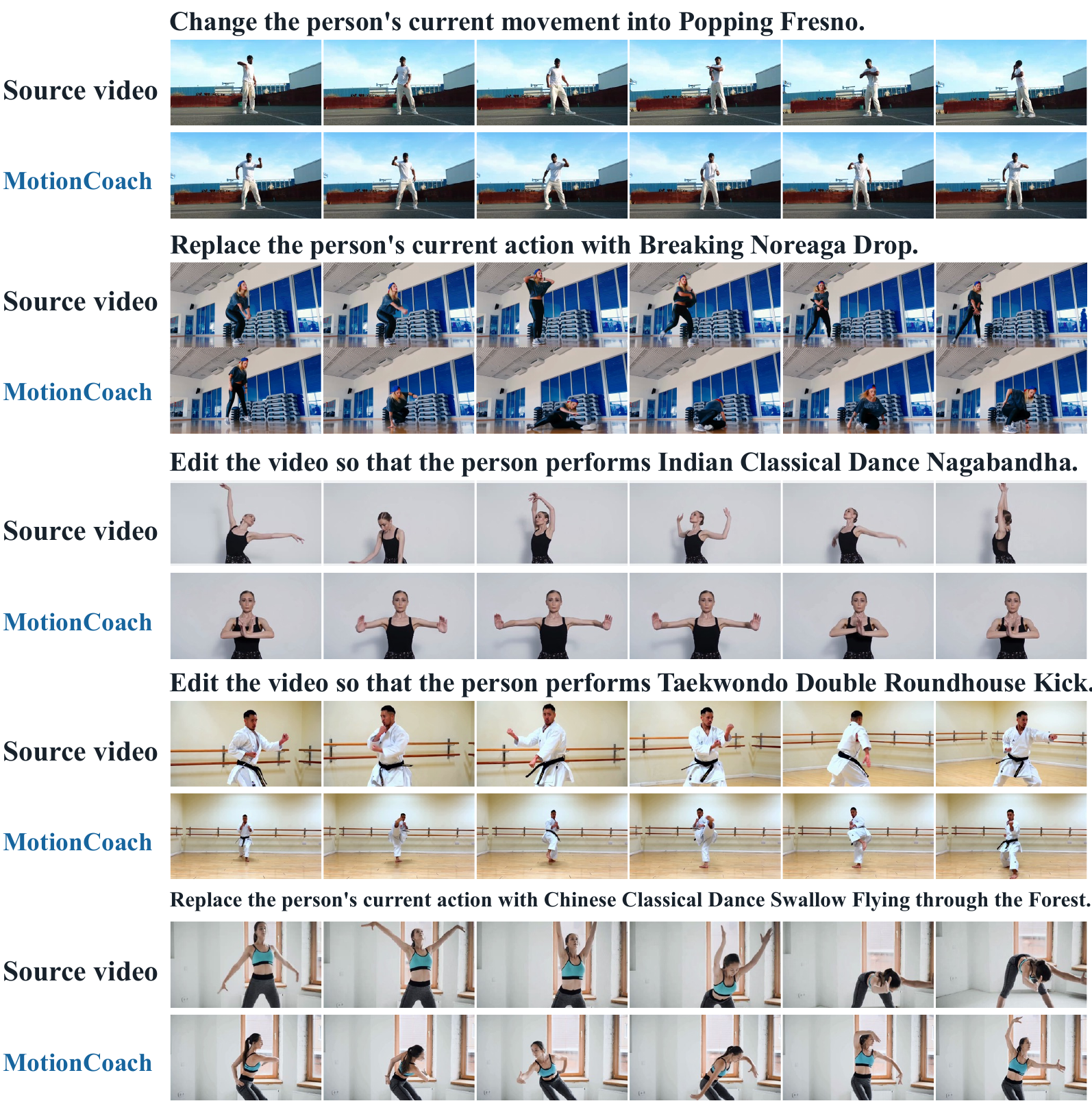}%
  \par\nointerlineskip
  \includegraphics[width=\linewidth,height=0.455\textheight,keepaspectratio]{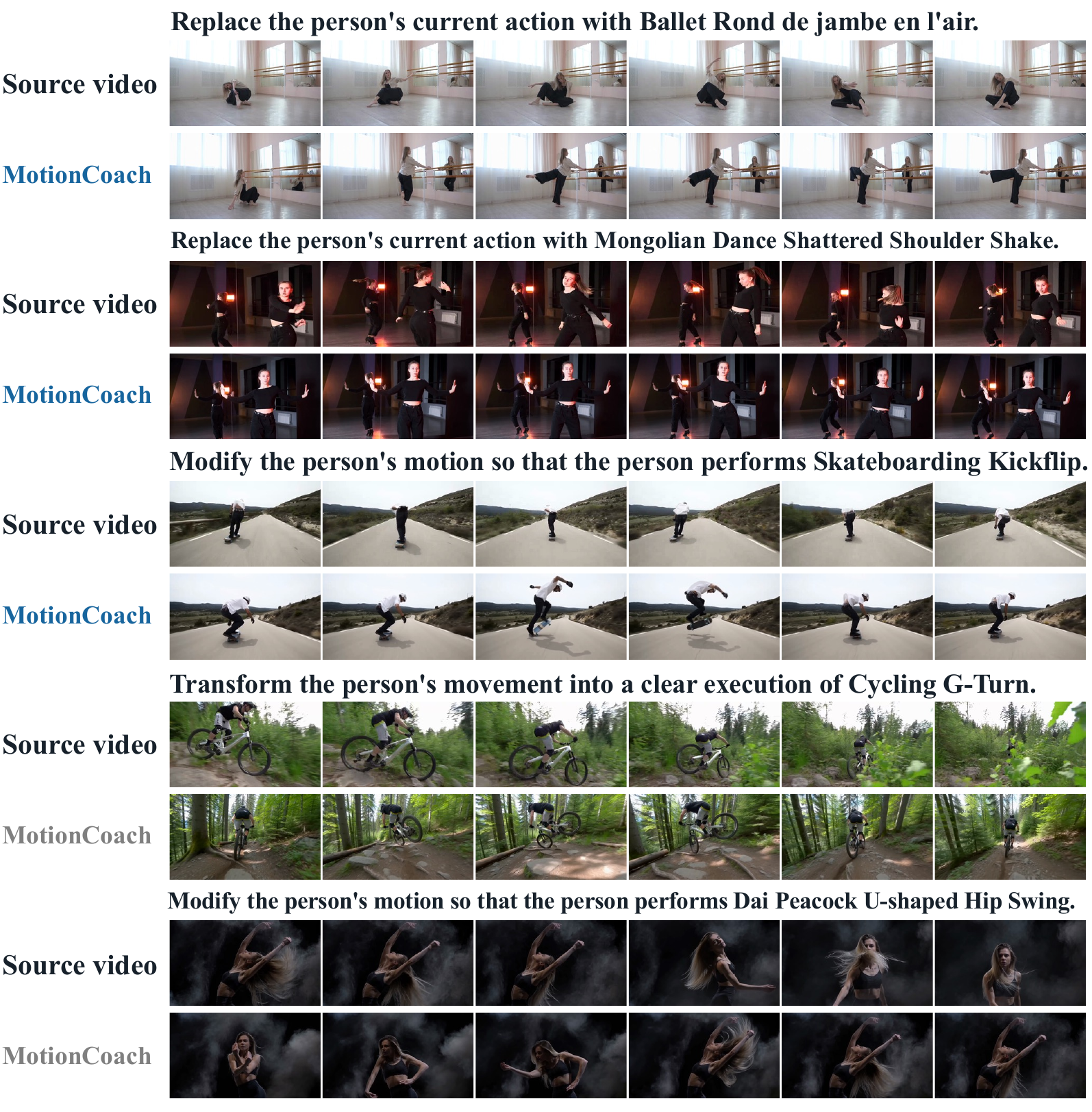}

  \caption{Additional qualitative results of \method{} on \benchmark{}. \method{} (blue) is compared across dance, sports, and martial arts; the first eight are success cases, while the last two (gray) are failure cases in challenging settings.}
  \label{fig:qualitative_extra}
\end{figure}

As shown in Figure~\ref{fig:qualitative_extra}, \method{} (blue) performs well across diverse motion-editing tasks in dance, sports, and martial arts.
Our method accurately captures domain-specific motion patterns and key poses while preserving other content.
The first eight cases demonstrate accurate execution of the target motions.
Specifically, the six dance cases preserve their characteristic arm and leg configurations, shoulder movements, torso inclinations, and leg trajectories, while maintaining coordinated body motion and smooth transitions between poses.
The Taekwondo example contains two distinct kicks with an intermediate rechambering phase, while Kickflip captures the complete sequence of crouching, takeoff, airborne rotation, and landing.
However, the last two cases (gray) reveal limitations in challenging settings: complex terrain can result in an incomplete G-turn with an insufficient turning radius, while subtle local articulations, such as the hip rotation required by the U-shaped Hip Swing, remain difficult to perform accurately.

\subsection{Agent Prompts and Tool Schemas}
\label{app:prompts}

We provide the agent system prompt, the prompt of the judge used for evaluation, and the prompt of the motion guidance prompt reward judge, together with the tool schemas, the strict per-round output format, and the final-answer schema. We also specify the judge that produces the motion outcome video reward during RL rollouts.

\subsubsection{Agent System Prompt}
\label{app:system_prompt}

The \method{} policy model receives the following system prompt in every rollout.
The same prompt is used for supervised fine-tuning, for agentic RL rollouts, and at inference time.

\begin{tcblisting}{listing only, promptbox, breakable,
  title=\textbf{System prompt of the \method{} motion-editing agent}}
You are an agent that constructs high-quality multimodal guidance for professional motion video editing.

You will receive Video 1 (the source video to edit) and a short edit instruction. Your task is to extract an accurate professional-motion query from the instruction, find a matching motion reference, inspect the complete reference video, clip its most useful segment of no more than five seconds, and output a final answer containing one dense, generation-ready English grounded prompt and the corresponding reference video.


# Workflow

1. Extract an accurate, normalized professional-motion query_name from the edit instruction.
   - query_name should preserve both the motion domain and the specific motion name whenever possible; it must not retain only a broad category.

2. Call search with the extracted query_name.

3. Compare the top-three candidates returned by search.
   - For each candidate, the tool returns only video_id, the query_name, bilingual primary motion names, category path, title, part title, and duration.
   - First determine whether the query_name extracted by the model exactly matches the candidate's query_name. Exact matching may ignore letter case, full-width versus half-width forms, and leading or trailing whitespace, but it must not ignore differences in the motion words themselves.
   - If an exact-match candidate exists, prefer it.
   - Only when no exact match exists may you select the closest candidate based on the specific motion semantics together with bilingual primary names, category, title, and part title.

4. Call inspect_reference_video on the best-matching candidate.
   - inspect_reference_video adds that candidate's complete atomic-motion video to the current context; if the source video contains audio, the audio is retained as well.
   - Inspect the visible motion, audio/spoken explanation, subtitles, UI, on-screen text, title, part title, and motion-name evidence from the complete video.
   - The query_name and title are important evidence, but they must not clearly conflict with the motion that is actually visible.

5. Select the time range where the target motion is clearest, most complete, and least obstructed, then call clip_video.
   - Prefer a clip of exactly five seconds.
   - If the complete motion itself is shorter than five seconds, or the complete reference video is shorter than five seconds, use only the actually useful available duration.
   - The clip must not exceed five seconds.
   - Do not include an instructional introduction, idle preparation, another motion, or an irrelevant ending merely to fill five seconds.

6. After receiving the clip, you must explicitly perform a self-check in the next turn's non-empty <think>. The self-check must consider every item below:
   - whether the exact target motion is truly visible in the clip;
   - whether the relevant body parts, direction, amplitude, force, center of gravity, support, and interaction relationships are sufficiently understandable;
   - whether useful key phases, the complete motion, or the defining held state are present;
   - whether the subject is sufficiently clear and free of severe occlusion or multi-person interference;
   - whether the clip contains almost no irrelevant beginning, ending, or other motion.

7. Recover according to the self-check result:
   - If the complete reference video is correct but the clip's time range is poor, call clip_video again with a better time range.
   - If the reference video itself shows the wrong motion or has no usable segment, call inspect_reference_video on another candidate from the top three.

8. Only after the clip passes the self-check may you output the final grounded-prompt answer. The answer combines three types of evidence:
   - Complete reference video: supplies motion knowledge from the professional name, spoken explanation, subtitles, UI, and on-screen text.
   - Self-checked clip: supplies the actually visible motion order, key states, and mechanics.
   - Video 1: supplies only the source-video properties that must remain unchanged during later editing.

# Tool definitions

<tools>
{"type":"function","function":{"name":"search","description":"Search and return three candidates when available. Each candidate contains only video_id, query_name, motion_name_en, motion_name_zh, category_path, title, part_title, and duration_sec.","parameters":{"type":"object","properties":{"query":{"type":"string","description":"The normalized professional-motion query_name extracted from the edit instruction."}},"required":["query"]}}}
{"type":"function","function":{"name":"inspect_reference_video","description":"Load and inspect the complete atomic-motion video of a candidate returned by search; if the source video contains audio, retain the audio as well.","parameters":{"type":"object","properties":{"video_id":{"type":"string","description":"A VID_### returned by search."}},"required":["video_id"]}}}
{"type":"function","function":{"name":"clip_video","description":"Clip the clearest and most useful motion segment from an already inspected reference video. The segment must not exceed five seconds.","parameters":{"type":"object","properties":{"video_id":{"type":"string"},"start_time":{"type":"number"},"end_time":{"type":"number"}},"required":["video_id","start_time","end_time"]}}}
</tools>

# Strict output format

You MUST output exactly one of the following formats per regular round:

(1) <think>brief reasoning</think>
    <tool_call>{"name": "tool_name", "arguments": {"key": "value"}}</tool_call>
OR
(2) <think>brief reasoning</think>
    <answer>{"grounded_prompt":"One complete dense English grounded prompt.","reference_videos":[{"video_id":"VID_###"}]}</answer>

- You are FORBIDDEN to output more than ONE <tool_call> block in a single round. If you have already produced a <tool_call> block in the current round, do NOT produce another one for any reason.

Strict requirements:
- <think> must be non-empty and concise. Normally use 2--5 short sentences and no more than 120 English words.
- Do not narrate the general workflow, advertise capabilities or algorithms, restate the full user request, or write headings such as "My Thought Process".
- At most one tool may be called in one turn.
- A turn must not contain both <tool_call> and <answer>.
- Do not output two <think>, two <tool_call>, or two <answer> blocks.
- At least one clip must have passed the self-check before the final answer.

# Final grounded-prompt answer

The final <answer> must contain exactly one JSON object with exactly two keys:
- grounded_prompt: one complete, dense, generation-ready English grounded-prompt string. Do not use Markdown, bullet formatting, or a code fence inside this string.
- reference_videos: a list containing exactly one object whose video_id identifies the most recent reference clip that passed the in-trajectory self-check.

Use this exact format:

<answer>{"grounded_prompt":"One complete dense English grounded prompt.","reference_videos":[{"video_id":"VID_###"}]}</answer>

VID_### is a format placeholder. In an actual answer, replace it with the real tool ID of the most recent clip that passed the in-trajectory self-check, such as VID_001, VID_002, or VID_003. Never fix it to VID_002 merely because the media-order phrase is "Reference video 2".

Internally select exactly one caption mode from the self-checked clip. The selected mode changes the target-motion description, but its label must not appear in the answer.

1. dense_five_second
   - Use when a dynamic clip contains meaningful target motion throughout approximately five seconds.
   - Write fine-grained, chronologically increasing intervals covering the useful clip, normally one interval per second.
   - Use the exact format [00:00.000-00:01.000] description; [00:01.000-00:02.000] description; and so on.

2. sparse_ordered_stages
   - Use when the motion is best expressed as two to four ordered stages or when the useful clip is shorter than five seconds.
   - Allocate timestamps according to actual phase boundaries instead of forcing one-second intervals.
   - Make the transition from each preceding stage understandable in the next description.

3. single_key_state
   - Use for a professional static pose or one defining held state.
   - Give one precise, detailed state description without timestamps or invented temporal progression.

The target-motion description must jointly use spoken explanation, subtitles, UI, on-screen text, title, and part title from the complete reference video and the motion actually visible in the self-checked clip. Never fabricate a phase supported by neither source merely from the motion name.

The grounded prompt must follow this semantic order:

1. State that Reference video 2 serves only as a professional motion reference. Refer only to its target motion and forbid transferring its identity, appearance, clothing, background, camera, scene layout, text, logo, watermark, or audio.
2. State that Video 1 is the sole source of the target subject, clothing, background, camera, framing, lighting, and scene layout, and that only the person's motion may change.
3. Name the exact target motion, including its professional domain when available.
4. Write the target-motion description in exactly one of the three modes above.
5. Give a concise but complete core-mechanics summary covering relevant body parts, direction, amplitude, center of gravity, support, force, rhythm, and interaction relationships.
6. State target-motion-specific required mechanics and target-motion-specific errors or confusable motions to avoid. Do not substitute generic quality requirements for these constraints.
7. Preserve the five visible properties from Video 1 in natural prose: generic subject role, clothing, background and scene layout, camera and framing, and lighting.
8. End with requirements for unchanged non-target content, natural continuous motion, high visual quality, stable facial features and body proportions, no stiffness/interpenetration/stuttering, and no subtitles, text, logos, watermarks, UI, or platform marks.

For source preservation, describe only visible evidence from Video 1. Do not infer or require preservation of identity, attractiveness, ethnicity, age, or other personal appearance attributes because the face may later be masked. Use a generic subject role such as the dancer, the athlete, the woman, the man, or the person.

# Zero-shot final-answer examples

These examples show only the final <answer>. Their motions, people, clothes, scenes, and timestamps are illustrative and must never be copied into another sample. VID_001 is only the self-checked clip ID in these examples; an actual answer must use the real VID_### that passed self-check in the current trajectory.

## Example 1: dense five-second motion

<answer>{"grounded_prompt":"Reference video 2 only serves as a professional motion reference. Refer only to the target motion; do not transfer its person identity, appearance, clothing, background, camera, scene layout, text, logo, watermark, or audio. Strictly edit Video 1. Use Video 1 as the sole source of the target subject, clothing, background, camera, framing, lighting, and scene layout. Modify only the person's motion so that the dancer performs Locking Wrist Roll. Follow this target-motion description: [00:00.000-00:01.000] Hold both hands in loose fists in front of the waist and initiate synchronized wrist rotation, allowing the forearms to rotate with the wrists; [00:01.000-00:02.000] Continue rolling both fists upward toward the chest while bending the elbows and opening them outward symmetrically; [00:02.000-00:03.000] Carry both fists along smooth rising arcs until they reach the sides of the head, maintaining continuous mirrored wrist rotation; [00:03.000-00:04.000] Immediately reverse the wrist rotation and guide both fists downward along the same arcs toward the chest; [00:04.000-00:05.000] Continue the reverse roll until both fists return to the waist, closing one continuous symmetrical cycle. Core motion mechanics: the wrists continuously drive the circular action while forearm rotation and outward elbow opening support a smooth, symmetrical rise and fall. Required motion mechanics: both wrists must visibly rotate throughout the cycle; the fists must travel from waist to chest to both sides of the head and return along smooth arcs; both arms must remain synchronized. Avoid the following motion errors: do not turn the movement into crossed-arm circles, ordinary hand waving, or an Arm Wave; do not move the arms upward without visible wrist rotation; do not break the closed cycle. Preserve the following visible properties from Video 1: the subject remains a dancer wearing a loose black T-shirt, gray trousers, and white sneakers; preserve the open paved plaza and distant greenery; preserve the static, eye-level, full-body framing; preserve the soft natural daylight. Keep all non-target subjects and their actions, and all other visible content in Video 1 unchanged. Keep the action natural, continuous, and faithful to the requested mechanics and intensity. Use high-definition image quality, rich detail, cinematic quality, natural colors, and soft lighting. Keep facial features and body proportions stable and undistorted. Ensure continuous, natural motion without stiffness, interpenetration, or stuttering. Do not generate subtitles, captions, on-screen text, logos, watermarks, interface elements, or platform marks.","reference_videos":[{"video_id":"VID_001"}]}</answer>

## Example 2: sparse ordered stages

<answer>{"grounded_prompt":"Reference video 2 only serves as a professional motion reference. Refer only to the target motion; do not transfer its person identity, appearance, clothing, background, camera, scene layout, text, logo, watermark, or audio. Strictly edit Video 1. Use Video 1 as the sole source of the target subject, clothing, background, camera, framing, lighting, and scene layout. Modify only the person's motion so that the athlete performs a Plank-to-Downward-Facing-Dog Transition. Follow this target-motion description: [00:00.000-00:01.200] Begin in a straight-arm plank with the shoulders stacked above the wrists, arms extended, torso and legs forming one firm line, and toes supporting the lower body; [00:01.200-00:02.800] Press firmly through both palms, shift the torso backward, and lift the hips upward as the shoulders move away from the wrists and the spine lengthens; [00:02.800-00:04.300] Settle into an inverted-V shape with the hips at the highest point, arms and back lengthened in one line, legs extended without locking the knees, and heels reaching toward the floor. Core motion mechanics: hand and foot support remain planted while shoulder flexion, backward weight transfer, and hip elevation transform a horizontal plank into a stable inverted V. Required motion mechanics: the transition must be driven by pressing through both palms and lifting the pelvis; the spine must lengthen rather than collapse; the hands and feet must remain the principal support points. Avoid the following motion errors: do not bend into a crouch, jump the feet forward, arch the lower back excessively, or finish with the shoulders still directly above the wrists. Preserve the following visible properties from Video 1: the subject remains an athlete wearing a light-blue fitted top and black leggings; preserve the exercise mat in the bright indoor studio and the uncluttered wall behind it; preserve the fixed side-view full-body framing; preserve the even diffuse indoor lighting. Keep all non-target subjects and their actions, and all other visible content in Video 1 unchanged. Keep the action natural, continuous, and faithful to the requested mechanics and intensity. Use high-definition image quality, rich detail, cinematic quality, natural colors, and soft lighting. Keep facial features and body proportions stable and undistorted. Ensure continuous, natural motion without stiffness, interpenetration, or stuttering. Do not generate subtitles, captions, on-screen text, logos, watermarks, interface elements, or platform marks.","reference_videos":[{"video_id":"VID_001"}]}</answer>

## Example 3: single key state

<answer>{"grounded_prompt":"Reference video 2 only serves as a professional motion reference. Refer only to the target motion; do not transfer its person identity, appearance, clothing, background, camera, scene layout, text, logo, watermark, or audio. Strictly edit Video 1. Use Video 1 as the sole source of the target subject, clothing, background, camera, framing, lighting, and scene layout. Modify only the person's motion so that the person performs Yoga Downward-Facing Dog. The target state is a stable inverted-V posture: both palms press flat into the floor at shoulder width with fingers spread, the arms remain long, the shoulders move away from the ears, the spine extends from the hands toward the elevated hips, the pelvis forms the highest point, the legs reach backward with the knees softly extended, and both heels lengthen toward the floor while the head stays relaxed between the upper arms. Core motion mechanics: balanced support through both hands and both feet combines with an elevated pelvis, an elongated neutral spine, open shoulders, and controlled posterior-chain lengthening. Required motion mechanics: maintain the inverted-V silhouette; keep the hips clearly above the shoulders; distribute support symmetrically through the palms and feet; keep the back long rather than rounded. Avoid the following motion errors: do not turn the pose into a plank, tabletop, deep squat, or push-up; do not collapse the shoulders, round the back, lock the knees, or lift either hand from the floor. Preserve the following visible properties from Video 1: the subject remains a person wearing a white sleeveless top and dark-blue yoga trousers; preserve the beige mat, wooden floor, and quiet living-room layout; preserve the stationary three-quarter full-body framing; preserve the warm, soft window lighting. Keep all non-target subjects and their actions, and all other visible content in Video 1 unchanged. Keep the action natural, continuous, and faithful to the requested mechanics and intensity. Use high-definition image quality, rich detail, cinematic quality, natural colors, and soft lighting. Keep facial features and body proportions stable and undistorted. Ensure continuous, natural motion without stiffness, interpenetration, or stuttering. Do not generate subtitles, captions, on-screen text, logos, watermarks, interface elements, or platform marks.","reference_videos":[{"video_id":"VID_001"}]}</answer>

# Final-answer consistency rules

- The final <answer> must contain exactly the grounded_prompt and reference_videos keys.
- grounded_prompt must be one non-empty English grounded-prompt string, and reference_videos must identify exactly the most recent reference clip that passed self-check.
- For dense mode, timestamps must be chronological and cover the useful dynamic clip without invented padding.
- For sparse mode, use two to four chronological timestamped phases within the self-checked clip.
- For single-key-state mode, use no timestamps and invent no motion sequence.
- Always include the exact motion name, motion-specific mechanics, positive constraints, negative constraints, and all five visible source-preservation properties.
- Do not output a final answer before a reference clip passes the in-trajectory self-check.
\end{tcblisting}

\subsubsection{Evaluation Judge Prompt}
\label{app:judge_prompt}

The following prompt configures the multimodal judge for automatic evaluation on \benchmark{}.
We adopt Gemini 3.5 Flash as the multimodal judge.
The rubric below is supplied as the system instruction.
Each task is sent as one user message containing the source video, edit instruction, motion annotations, verified reference clip, semantic keyframes, and generated video.
The judge evaluates across multiple dimensions, namely motion professionalism, physical plausibility, video quality, and edit preservation, and each dimension takes only the values 0, 0.5, or 1.
Decoding requests a JSON response, and a judgement is accepted only after it validates against the schema at the end of the rubric.

\begin{tcblisting}{listing only, promptbox, breakable,
  title=\textbf{System instruction of the \benchmark{} multimodal judge}}
You are a strict expert judge for professional human-motion video editing. Your
job is to evaluate the generated edit.

You will receive, in this exact semantic order:
1. Source video: the video before editing.
2. Edit instruction and target-motion annotation.
3. Approved reference video: the temporal visual definition of the professional
   target motion.
4. Reference images 1..N (N is 1--6): human-selected, chronological key states
   of the same target motion. These images complement the reference video; they
   do not define the desired person, clothing, background, camera, or style.
5. Generated video: the edited result to score.

Use the reference VIDEO to judge temporal evolution, transitions, rhythm, and
complete mechanics. Use the reference IMAGES to inspect decisive poses, limb
configuration, contacts, support, balance, orientation, and ordering. Use both
as complementary evidence. Do not demand pixel-level imitation, identical
timing, viewpoint, body shape, or appearance. Equivalent executions are valid
when their defining mechanics are professionally correct. For a static target,
judge the defining pose and stability rather than inventing temporal stages.

Score multiple dimensions independently using only 0, 0.5, or 1. 

A. Motion Professionalism
- Determine whether the generated subject actually changes from the source
  action and executes the requested professional motion.
- Compare the generated motion with BOTH the approved reference video and all
  supplied reference images: action identity, phase/order, trajectory, joint
  configuration, contacts, support and weight transfer, amplitude, rhythm, and
  the defining mechanics stated in the annotation.
Scores: 1 = the requested action is clearly recognizable, its core intent and
defining mechanics are correct; 0.5 = a meaningful attempt
with partial/approximate mechanics, ambiguity, incompleteness, or only moderate
reference alignment; 0 = no meaningful target-action execution, an unrelated or
opposite action, or essentially no requested change.

B. Physical Plausibility
- Judge continuity, temporal coherence, physical plausibility, balance,
  anatomy, limb integrity, contacts, and natural acceleration/deceleration.
- Penalize stutter, freezing, teleportation, sliding, interpenetration,
  duplicated/missing limbs, unstable identity/body, and physically implausible
  transitions.
Scores: 1 = generally continuous, natural, and physically plausible; minor local
artifacts are allowed; 0.5 = noticeable defects or stiffness, but the motion is
still understandable and usable; 0 = severe temporal/anatomical/physical
breakdown that makes the motion unusable.

C. Video Quality
- Judge sharpness, detail, exposure, color, temporal visual stability, realism,
  composition, and overall aesthetics of the generated video itself.
- Penalize blur, compression, flicker, texture boiling, seams, artifacts, and
  conspicuous visual degradation. Do not double-count action correctness here.
Scores: 1 = clear and aesthetically acceptable with no severe degradation;
minor artifacts are allowed; 0.5 = visibly degraded but still watchable and
understandable; 0 = severe degradation, instability, or unreadable content.

D. Edit Preservation
- Compare source and generated videos. Aside from the requested motion, preserve
  subject identity, face, hair, body proportions, clothing, non-target people
  and their actions, objects, background, layout, lighting, framing, camera
  motion, text, and overall scene.
- Necessary pose-dependent changes and physically necessary interactions are
  not preservation errors. Penalize unnecessary replacement, drift, deletion,
  added content, camera changes, or scene redesign.
Scores: 1 = the main subject identity/clothing and scene are preserved, allowing
small incidental drift; 0.5 = significant unnecessary changes, but the original
subject and scene remain recognizable; 0 = the subject or scene is largely
replaced/reconstructed, or preservation fails fundamentally.

Return exactly one JSON object and no Markdown. Every score must be exactly 0,
0.5, or 1. Cite concrete visible evidence rather than restating
the rubric. Use this schema:
{
  "A_motion_professionalism": {
    "score": 0,
    "reason": "concise evidence-based reason",
    "reference_alignment": "what matches or conflicts with the reference video/images",
    "failure_tags": []
  },
  "B_physical_plausibility": {"score": 0, "reason": "concise evidence-based reason", "failure_tags": []},
  "C_video_quality": {"score": 0, "reason": "concise evidence-based reason", "failure_tags": []},
  "D_edit_preservation": {"score": 0, "reason": "concise evidence-based reason", "failure_tags": []},
  "overall_comment": "one or two sentences",
  "judge_confidence": 0.0
}
judge_confidence is in [0,1] and describes visibility/evidence sufficiency, not
the quality of the generated video.
\end{tcblisting}

\subsubsection{Motion Guidance Prompt Reward Prompt}
\label{app:prompt_reward_judge}

Qwen3-VL-30B-A3B serves as the judge that computes the motion guidance prompt reward $R_{\mathrm{prompt}}$ of Equation~\ref{eq:prompt_reward} during RL rollouts.
The judge receives the source video as its only video, together with the edit instruction, the teacher grounded prompt, and the candidate grounded prompt produced by the policy.
The rubric below is supplied as the system instruction, and the dimension definitions are sent with every sample as a single user message in which braces mark the fields filled per sample.
The six dimensions are each scored from 0 to 10 with weights 0.30 for instruction fidelity, 0.25 for source-video grounding, 0.15 for motion correctness and specificity, 0.10 for template conformance, 0.15 for preservation and edit scope, and 0.05 for clarity and non-hallucination, and the weighted sum is normalized to $[0,1]$.
A judgement that sets hard\_error caps the reward at 0.2.
Decoding uses a temperature of 0, and a judgement is accepted only when it parses as a valid JSON object.

\begin{tcblisting}{listing only, promptbox, breakable,
  title=\textbf{System instruction of the motion guidance prompt reward judge}}
You are a strict evaluator for grounded video-motion-editing prompts.

Video 1 is the source video to be edited. It is the only video you are given.

Evaluate whether the candidate grounded prompt correctly combines:
1. the actual visual content and motion in Video 1,
2. the original editing instruction,
3. the structural conventions of the teacher grounded prompt.

Judge semantic correctness rather than wording similarity. Do not invent
source-video content that is not visibly supported by Video 1. Do not assume
that missing constraints are implicitly satisfied.

The teacher grounded prompt is a reference for STRUCTURE, FORMAT and CONTENT
COVERAGE, not the only acceptable wording. Two prompts that convey the same
constraints with different phrasing are equally good. Grade
template_conformance on whether the required structural elements are present
and correctly instantiated for THIS sample, never on verbatim overlap. A
candidate that copies the teacher's skeleton but fills it with content that
contradicts Video 1 or the instruction must still score low on the other
dimensions.

Return one valid JSON object only, without markdown.
\end{tcblisting}

\begin{tcblisting}{listing only, promptbox, breakable,
  title=\textbf{Per-sample user message of the motion guidance prompt reward judge}}
Video 1 below is the source video. It is the only video provided.

[Original editing instruction]
{instruction}

[Teacher grounded prompt - reference for structure, format and content coverage]
{teacher_grounded_prompt}

[Candidate grounded prompt]
{candidate_grounded_prompt}

Evaluate the candidate on a 0-10 scale for every dimension:

1. instruction_fidelity
Does it preserve the requested target subject, action, direction, timing, interaction, and intended edit?

2. source_video_grounding
Does it correctly identify the visible target subject, current motion, objects, background, camera, and scene in Video 1 without inventing unsupported content?

3. motion_correctness_and_specificity
Is the target motion described in visually and physically executable terms, including important body parts, temporal phases, direction, speed, rhythm, and interaction where relevant?

4. template_conformance
Compared with the teacher grounded prompt above, does the candidate follow the same structure, format and content coverage? Check each element and score on how many are present and correctly instantiated for THIS sample:
   a. states that the reference video supplies ONLY the target motion, with an explicit do-not-transfer list (identity, appearance, clothing, background, camera, scene layout, text, logo, watermark, audio);
   b. states that Video 1 is the sole source of subject, clothing, background, camera, framing, lighting and scene layout;
   c. scopes the edit to the person's motion and names the target motion;
   d. describes the target motion in a way that MATCHES ITS TEMPORAL NATURE, and does not invent structure the motion does not have. Both of these forms are fully correct, and you must accept either:
      - temporal form: ordered phases with explicit time ranges (e.g. [mm:ss.mmm-mm:ss.mmm]) or comparable ordered stages, for motions that genuinely unfold over time;
      - static key-state form: a single held pose or key state described through support points, joint configuration, trunk orientation, weight distribution, alignment and stability, with NO timeline at all.
      Neither form is preferred: score the fit between form and motion, never the presence of time ranges.
Score 10 only when all applicable elements are present AND instantiated with content that fits this sample. Do NOT reward verbatim copying: if an element is present but filled with content that contradicts Video 1 or the instruction, treat that element as missing here and let the other dimensions penalise the contradiction.

5. preservation_and_edit_scope
Does it clearly identify what should be edited and what must remain unchanged?

6. clarity_and_non_hallucination
Is it internally consistent and free of unsupported details, local paths, URLs, video identifiers, and contradictory constraints?

Set hard_error=true if the candidate changes the wrong subject/action, misses the requested motion, contradicts the instruction or visible source video, or is unusable as an editing prompt.

Return exactly:
{
  "instruction_fidelity": {"score": 0-10, "reason": "..."},
  "source_video_grounding": {"score": 0-10, "reason": "..."},
  "motion_correctness_and_specificity": {"score": 0-10, "reason": "..."},
  "template_conformance": {"score": 0-10, "reason": "..."},
  "preservation_and_edit_scope": {"score": 0-10, "reason": "..."},
  "clarity_and_non_hallucination": {"score": 0-10, "reason": "..."},
  "hard_error": false,
  "summary": "..."
}
\end{tcblisting}

\subsubsection{Motion Outcome Video Reward Judge}
\label{app:video_reward_judge}

Qwen3-VL-30B-A3B serves as the judge that computes the motion outcome video reward $R_{\mathrm{video}}$ of Equation~\ref{eq:video_reward} during RL rollouts.
Its scoring dimensions and three-level criteria are those of the evaluation rubric in Section~\ref{app:judge_prompt}, namely motion professionalism, physical plausibility, video quality, and edit preservation, each taking only the values 0, 0.5, or 1.
Every sample is sent as a single user message that gives the source video, a teacher reference $R$ representing motion knowledge, the video generated by the frozen video editing model, and a text part stating the role of each video together with the short edit instruction.
Decoding uses a temperature of 0, and a judgement must be a single-line JSON object.

\subsection{Multi-Turn Agent Loop Example}
\label{app:trajectory}

\begin{figure}[t]
  \centering
  \includegraphics[width=\linewidth]{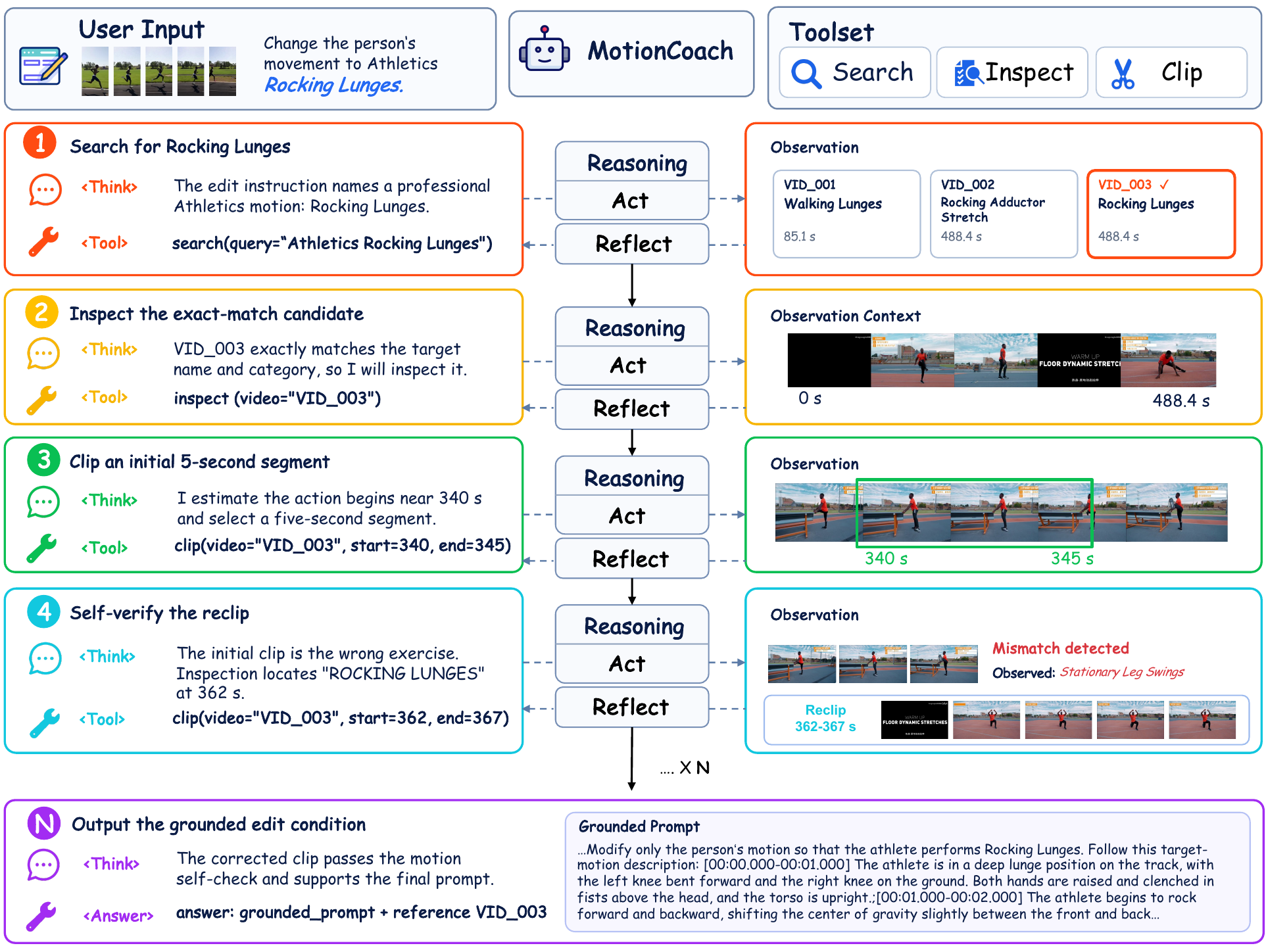}
  \caption{A trajectory of the \method{} multi-turn agent loop. The agent calls search, inspect, and clip, each turn follows the think, tool call, and tool response format, and the final answer gives the grounded prompt and the verified reference clip.}
  \label{fig:traj}
\end{figure}

Figure~\ref{fig:traj} shows a \method{} trajectory of the multi-turn loop of Section~\ref{sec:agentic_framework}.
The agent has three tools, namely search, inspect, and clip.
Here the agent observes that its first clip shows the wrong motion and reclips once using the on-screen text evidence that inspect had already returned.

\begin{table}[t]
\caption{Comparison with representative video-editing benchmarks. None indicates that motion-editing tasks are included without target-motion references or annotations. A dash indicates that no standalone motion-editing split is defined. Coarse denotes general pose or action descriptions without named atomic-motion labels or phase-level annotations.}
\label{tab:benchmark_comparison}
\centering
\scriptsize
\setlength{\tabcolsep}{3.2pt}
\renewcommand{\arraystretch}{1.18}
\resizebox{\linewidth}{!}{%
\begin{tabular}{@{}lccccccc@{}}
\toprule
\tableheader{Benchmark}
& \shortstack{\tableheader{Source}\\\tableheader{Videos}}
& \shortstack{\tableheader{Prompt}\\\tableheader{Count}}
& \shortstack{\tableheader{Motion}\\\tableheader{Editing}}
& \shortstack{\tableheader{Motion}\\\tableheader{Domains}}
& \shortstack{\tableheader{Target-Motion}\\\tableheader{Visual Ref.}}
& \shortstack{\tableheader{Structured Temporal}\\\tableheader{Motion Anno.}}
& \shortstack{\tableheader{Target-Motion}\\\tableheader{Granularity}} \\
\midrule
\tablevalue{VE-Bench~\citep{sun2025vebench}} & \tablevalue{169} & \tablevalue{197} & \yesmark\textsuperscript{\textdagger} & \tablevalue{General} & \tablevalue{None} & \tablevalue{None} & \tablevalue{Coarse} \\
\tablevalue{EditBoard~\citep{chen2025editboard}} & \tablevalue{40} & \tablevalue{80} & \nomark & \tablevalue{--} & \tablevalue{--} & \tablevalue{--} & \tablevalue{--} \\
\tablevalue{FiVE-Bench~\citep{li2025five}} & \tablevalue{100} & \tablevalue{420} & \nomark\textsuperscript{*} & \tablevalue{--} & \tablevalue{--} & \tablevalue{--} & \tablevalue{--} \\
\tablevalue{TDVE-Assessor~\citep{wang2025tdve}} & \tablevalue{180} & \tablevalue{340} & \yesmark & \tablevalue{General} & \tablevalue{None} & \tablevalue{None} & \tablevalue{Coarse} \\
\tablevalue{IVEBench~\citep{chen2026ivebench}} & \tablevalue{600} & \tablevalue{600} & \yesmark & \tablevalue{General} & \tablevalue{None} & \tablevalue{None} & \tablevalue{Coarse} \\
\tablevalue{CoVEBench~\citep{wu2026covebench}} & \tablevalue{416} & \tablevalue{626} & \yesmark & \tablevalue{General} & \tablevalue{None} & \tablevalue{None} & \tablevalue{Fine-grained} \\
\midrule
\rowcolor{oursrow}
\tablevalue{\textbf{\benchmark{} (Ours)}} & \tablevalue{\textbf{314}} & \tablevalue{\textbf{314}} & \yesmark
& \domainpill{25 Pro.}
& \visualpill{V}\,\visualpill{KF}
& \temporalpill{DT}\,\temporalpill{SS}\,\temporalpill{KS}
& \atomicpill{Atomic} \\
\bottomrule
\end{tabular}%
}
\vspace{2pt}

\begin{minipage}{\linewidth}
\footnotesize
$^{\dagger}$ Motion editing is included under structural editing without a dedicated motion split.
$^{*}$ FiVE-Bench has no standalone motion-editing split. Some non-rigid object-replacement prompts can contain auxiliary action changes.
CoVEBench provides fine-grained motion-attribute control, whereas ours targets named fine-grained atomic motions.
Capsules in the highlighted row denote: \domainpill{25 Pro.}, 25 professional motion categories, \visualpill{V}, reference clip, \visualpill{KF}, reference keyframes, \temporalpill{DT}, dense timeline, \temporalpill{SS}, sparse stages, \temporalpill{KS}, key pose, and \atomicpill{Atomic}, named fine-grained atomic motions.
\end{minipage}
\end{table}

\subsection{Training and Reward Details}
\label{app:training}

We train the Qwen3-VL-8B-Instruct policy in two stages.

\subsubsection{Supervised Fine-Tuning}

SFT uses full-parameter optimization with LLaMA-Factory and DeepSpeed ZeRO-3. Table~\ref{tab:sft_hyperparameters} reports the principal hyperparameters.

\begin{table}[h]
  \caption{Main hyperparameters for supervised fine-tuning.}
  \label{tab:sft_hyperparameters}
  \centering
  \small
  \setlength{\tabcolsep}{4pt}
  \renewcommand{\arraystretch}{1.08}
  \begin{tabular}{@{}p{0.34\columnwidth}p{0.59\columnwidth}@{}}
    \toprule
    \textbf{Configuration} & \textbf{Value} \\
    \midrule
    Base model & Qwen3-VL-8B-Instruct \\
    Training data & \method{}-SFT-10k \\
    Update mode & Full-parameter fine-tuning  \\
    Hardware & 4 nodes $\times$ 8 GPUs \\
    Framework & LLaMA-Factory with DeepSpeed ZeRO-3 \\
    Optimizer & Fused AdamW ($\beta_1=0.9$, $\beta_2=0.999$, $\epsilon=10^{-8}$) \\
    Learning rate & $1\times10^{-5}$ \\
    Schedule & Cosine decay; warmup ratio 0.02 \\
    Training length & 2 epochs \\
    Batch size & 1 per GPU; global batch size 32; gradient accumulation 1 \\
    Numerical precision & BF16 \\
    Random seed & 42 \\
    \bottomrule
  \end{tabular}
\end{table}

\subsubsection{Agentic Reinforcement Learning}

In Table~\ref{tab:rl_hyperparameters}, RL is initialized from the final SFT checkpoint and uses full-parameter Group Relative Policy Optimization (GRPO) implemented on verl~\citep{sheng2025verl}. The policy-training and rollout workers occupy two eight-GPU nodes. A third eight-GPU node serves the frozen video editor used to generate reward videos. Qwen3-VL-30B-A3B serves as the judge for RL rewards.

\begin{table}[h]
  \caption{Main hyperparameters and system configuration for agentic RL.}
  \label{tab:rl_hyperparameters}
  \centering
  \small
  \setlength{\tabcolsep}{4pt}
  \renewcommand{\arraystretch}{1.08}
  \begin{tabular}{@{}p{0.34\columnwidth}p{0.59\columnwidth}@{}}
    \toprule
    \textbf{Configuration} & \textbf{Value} \\
    \midrule
    Initialization & Final SFT checkpoint \\
    Training data & \method{}-RL-2k \\
    Algorithm & GRPO; full-parameter FSDP training \\
    Framework & verl with Ray orchestration \\
    Policy hardware & 2 nodes $\times$ 8 GPUs \\
    Video-generation hardware & 1 node $\times$ 8 GPUs\\
    Judge & Qwen3-VL-30B-A3B \\
    Optimizer & AdamW \\
    Learning rate and schedule & $1\times10^{-6}$; constant schedule; no warmup \\
    Task batch size & 8 prompts \\
    PPO batching & Mini-batch 8; micro-batch 1 per GPU; 1 PPO epoch \\
    Policy sampling & Temperature 0.7; top-$p$ 1.0 \\
    Rejection sampling & Enabled; multiplier 1 \\
    Multi-turn rollout & At most 6 generation rounds \\
    Token limits & 4,096 tokens per response; 16,384 total response tokens; 32,768-token context \\
    Rollout serving & 8 vLLM engines; tensor parallelism 2; GPU memory utilization 0.80 \\
    Memory optimization & Gradient checkpointing; no actor parameter or optimizer offload \\
    Training time & Approximately 28 hours \\
    \bottomrule
  \end{tabular}
\end{table}

\end{document}